\documentclass[a4paper,conference]{IEEEtran}
\usepackage{commath}

\usepackage{hyperref}

\ifCLASSINFOpdf
  \usepackage[pdftex]{graphicx}
  \graphicspath{{images/}}
  \DeclareGraphicsExtensions{.pdf,.jpeg,.png}
\else
\fi
\begin{document}
%
\title{FC-DCNN: A densely connected neural network for stereo estimation}

\author{\IEEEauthorblockN{Dominik Hirner}
\IEEEauthorblockA{Institute for Computer Graphics and Vision\\
Graz University of Technology\\
Austria\\
Email: dominik.hirner@icg.tugraz.at}
\and
\IEEEauthorblockN{Friedrich Fraundorfer }
\IEEEauthorblockA{Institute for Computer Graphics and Vision\\
Graz University of Technology\\
Austria\\
Email: fraundorfer@icg.tugraz.at}}


%



\maketitle
\begin{abstract}
We propose a novel lightweight network for stereo estimation. Our network consists of a fully-convolutional densely connected neural network (FC-DCNN) that computes matching costs between rectified image pairs. Our FC-DCNN method learns expressive features and performs some simple but effective post-processing steps. The densely connected layer structure connects the output of each layer to the input of each subsequent layer. This network structure and the fact that we do not use any fully-connected layers or 3D convolutions leads to a very lightweight network. The output of this network is used in order to calculate matching costs and create a cost-volume. Instead of using time and memory-inefficient cost-aggregation methods such as semi-global matching or conditional random fields in order to improve the result, we rely on filtering techniques, namely median filter and guided filter. By computing a left-right consistency check we get rid of inconsistent values. Afterwards we use a watershed foreground-background segmentation on the disparity image with removed inconsistencies. This mask is then used to refine the final prediction. We show that our method works well for both challenging indoor and outdoor scenes by evaluating it on the Middlebury, KITTI and ETH3D benchmarks respectively. Our full framework is available at \href{https://github.com/thedodo/FC-DCNN}{https://github.com/thedodo/FC-DCNN}

\end{abstract}


%
\IEEEpeerreviewmaketitle


\section{Introduction}

Retrieving 3D information from image pairs is a major topic in computer vision and has become even more popular in recent years because of the advances in autonomous driving, robotics and remote sensing.

A typical stereo method consists of the following four steps: feature extraction, matching cost calculation, disparity estimation and disparity refinement.
In the past, handcrafted feature extraction methods like Census~\cite{feat:census} or dense gradient features~\cite{feat:dense} were used. In recent years however, many applications have shown that deep learning methods are advantageous~\cite{disp:mc_cnn}\cite{disp:cnn_crf}\cite{disp:ga_net}\cite{disp:psm_net}\cite{disp:gc_net}\cite{disp:efficient_stereo} and can improve matching results in many real world challenges by learning. Such challenges for example include textureless areas like floors, walls or the sky, specular reflections on smooth surfaces or thin structures and clutter. 
By learning more expressive features for such challenging areas using deep learning techniques, the number of correct matches found can be improved. 

\begin{figure}[!t]
\centering
\includegraphics[width=9cm]{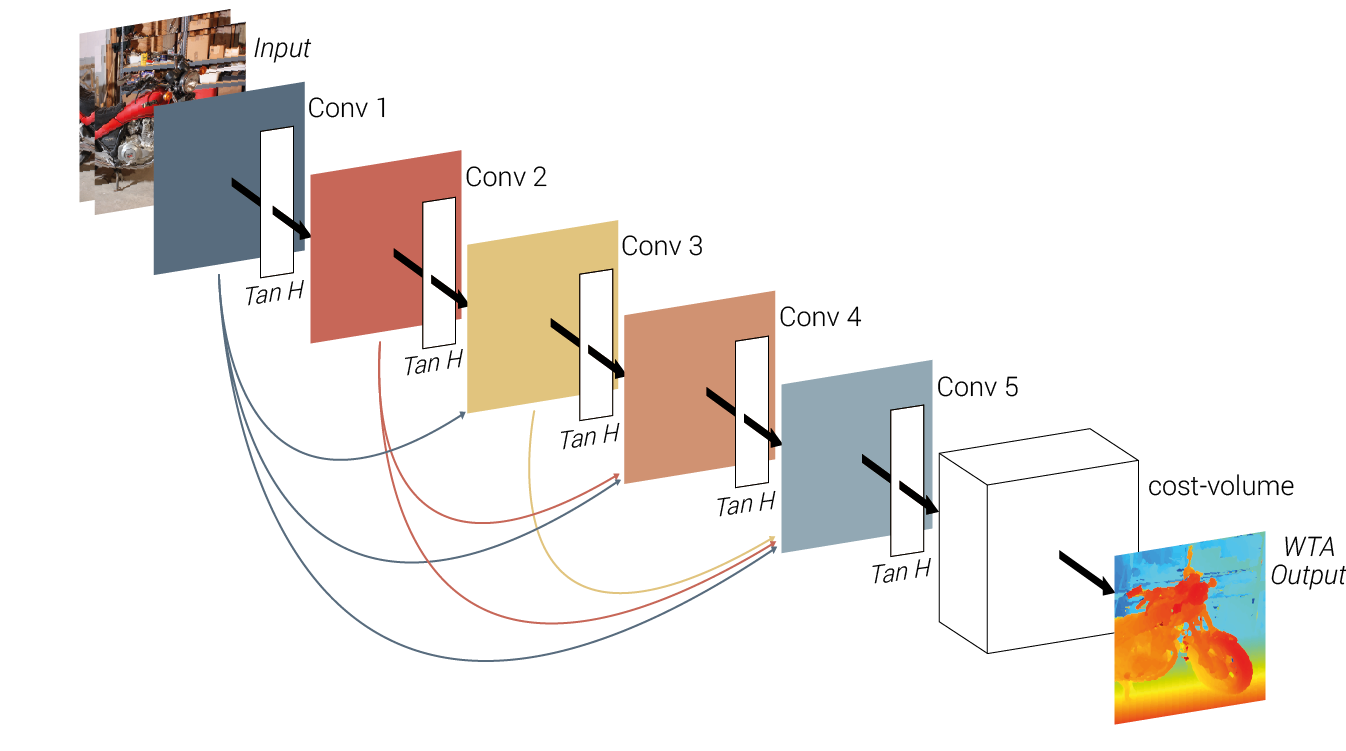}
\caption{FC-DCNN network structure. The left and the right image are processed at the same time by individual branches with shared weights. Each branch consists of 5 convolutional layers (Conv1-Conv5) with $k=3x3$ kernels and $m=64$ feature maps. After the last layer a cost volume is created using the cosine similarity. The final prediction is the winner-takes-all estimate along the third dimension of the cost-volume.}
\label{fig:network_arch}
\end{figure}

We follow the work of Zbontar and LeCun~\cite{disp:mc_cnn} by creating a shared-weights siamese network structure for feature extraction. However instead of using one connection between subsequent layers, we use a densely-connected network structure. As described by G. Huang et al.~\cite{densenet}, this structure helps to alleviate the vanishing-gradient problem due to better feature reuse and better feature propagation. This allows us to reduce the number of trainable parameters in comparison to traditional feed-forward CNN networks. Our whole network structure is illustrated in  Fig.~\ref{fig:network_arch}. The arrows in color depict the additional connections of the dense network structure between layers.

Matching costs are based on the similarity measurements between the extracted features of the image pairs. Over the years many different similarity measurements have been studied and proposed, such as the sum of absolute difference/sum of squared difference (SAD/SSD)~\cite{feat:sadssd}, normalized cross correlation (NCC)~\cite{feat:ncc} or Mutual Information (MI)~\cite{mi}. Previous works have integrated the learning of a matching cost function within their network architecture. The advantage of this is that the whole matching pipeline is fully automated and trainable, however we decided against it in our implementation as the corresponding parts of the network would highly increase the complexity while only improving slightly over traditional matching costs like normalized cross correlation~\cite{feat:ncc} or cosine similarity (in our experiments only about 1-2\%).

\begin{figure*}[!t]
\centering
\includegraphics[width=17cm]{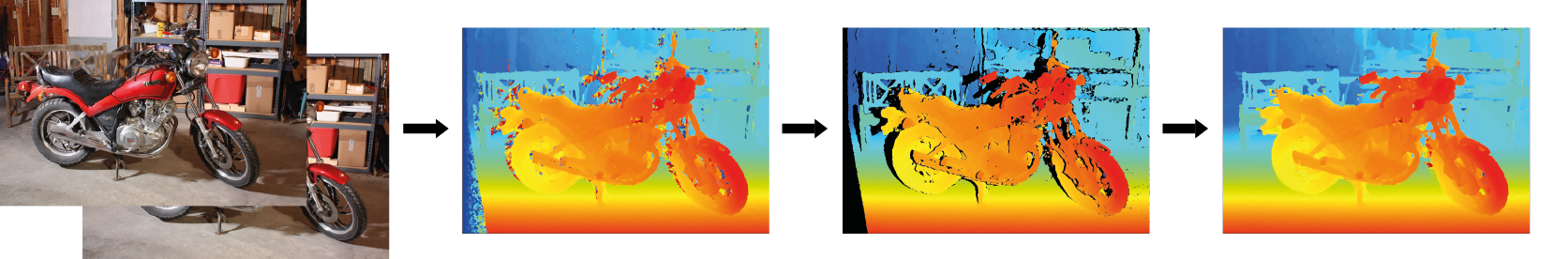}
\caption{All intermediate results of our method. From left to right: Input RGB stereo pair, winner-takes-all (WTA) output of the network, WTA output with removed inconsistencies, final prediction by filling in the previously removed values.}
\label{fig:all_results}
\end{figure*}

Once all the matches are calculated, the most likely candidate for each position is chosen. Even with better and more expressive learned features, the resulting disparity map can often still be subject to strong outliers and noise. This is why a post-processing or regularization step of the cost-volume is important. Traditional methods use regularization techniques such as semi-global matching (SGM)~\cite{reg:sgm}  or more-global matching (MGM)~\cite{reg:mgm}. However these regularization techniques, while still being competitive in regards to accuracy, are relatively slow and memory-inefficient because most implementations are not optimized for GPU usage. Therefore we use a pytorch implementation of the median filter~\cite{kornia} and guided filter~\cite{guided_filter} on each slice of the cost volume before taking the maximum for the final prediction instead. Afterwards we rely on a left-right consistency check to identify inconsistent points and use a watershed foreground-background segmentation in order to decide how to update these values. 
The input, all intermediate results and the final disparity estimation can be seen in Fig.~\ref{fig:all_results}.\\
In summary our contributions are as follows:
  \begin{itemize}
     \item We propose a novel fully-convolutional densely connected siamese network structure for feature extraction. We use dense-layer connections and do not use any fully-connected layers or 3D convolutions. Therefore we are able to produce a lightweight network structure.
     \item  We train and evaluate our network on three challenging datasets, namely Middlebury, KITTI and ETH3D. We discuss the results both qualitatively as well as quantitatively. We show that our method can compete with state-of-the-art methods.
     \item We implement our own post-processing based on filtering, finding inconsistencies via a left-right check and updating found inconsistent values by using a watershed algorithm on the disparity map with removed inconsistencies. This allows us to be independent from out-of-the box regularization techniques which might not be optimized for GPU usage.
  \end{itemize}

Our method can be seen as a hybrid method, which is faster and more accurate as traditional non-learning methods such as SGM~\cite{reg:sgm} while needing less GPU-Ressources than fully end-to-end methods such as PSMNet~\cite{psmnet} while still producing comparable accuracies and being well suited for typical applications.

\section{Related Work}

Our work is based on previous work on feature extraction using CNNs, similarity measurements and disparity refinement.

\textbf{Feature extraction} is an important step for any stereo method. While there are many still used handcrafted feature extractors such as Census~\cite{feat:census} or dense gradient features~\cite{feat:dense}, many new approaches use CNNs in order to learn more expressive and robust features. 
A popular model for this task is a siamese CNN structure with shared weights~\cite{disp:mc_cnn}\cite{disp:cnn_crf}\cite{disp:ga_net}\cite{disp:psm_net}\cite{disp:gc_net}\cite{disp:efficient_stereo}. This network structure was popularized by the work of Zbontar and LeCun~\cite{disp:mc_cnn} for the task of disparity estimation. In their work they extract small image patches from the left image and corresponding correct and wrong patches from the right image. They then formulate the training of the feature extractor as a binary classification where they want to maximize the distance in similarity between the correct and not correct pair of patches.

\textbf{Similarity measurements} are widely used in machine learning and are often a vital part of the loss function and training. In stereo vision similarity measurements are additionally used as the matching cost function in order to find corresponding points between the image pair. Many of the commonly used cost functions are window-based, as single pixel values are often not expressive enough to confidently find the correct match. Such window-based function include the sum of absolute difference/sum of squared difference (SAD/SSD)~\cite{feat:sadssd}, normalized cross correlation (NCC)~\cite{feat:ncc}, Census or Rank~\cite{feat:census}. H. Hirschmueller and D. Scharstein evaluated all previously mentioned matching costs and more in their paper~\cite{feat:eval_cost}.

Window-based matching costs however can be time- and ressource-inefficient if naively implemented due to the sliding window problem (though some improvements have been suggested~\cite{fast-rcnn}\cite{locnet}). We argue that due to the nature of deep-CNNs, each image point has already implicit knowledge of its immediate neighbourhood encoded in its multi-dimensional feature vector. We therefore use the pixel-wise cosine similarity as our cost function.

Many stereo networks do not use handcrafted cost functions but rather learn it together with the feature extractor as part of the network~\cite{disp:mc_cnn}\cite{disp:cnn_crf}, however most use fully-connected layers (or 1x1 convolution layers) in order to accomplish this and therefore increase their model complexity manifold. While we do not deny that there are advantages of a fully end-to-end trainable model, we decided against learning a cost function in order to keep our network smaller.

\textbf{Disparity refinement} is done in order to create the final disparity prediction. In this step the often still noisy and outlier/peak-prone output is taken and optimized. One of the most-popular methods for disparity refinement is semi-global matching (SGM)~\cite{reg:sgm}. In the original paper, Hirschmueller uses Mutual Information~\cite{mi} for the matching cost, this however can be substituted for any matching cost.
SGM aggregates the matching costs from all 16 cardinal direction for each pixel, by approximating 2D smoothness constraint by combining many 1D line optimization problems. 

G. Facciolo et al.~\cite{reg:mgm} improve upon this method by using different elements and using more than one cardinal direction for the belief update of one disparity value. He also discusses the artefacts that can be produced by the update scheme of SGM and its variants.

F. Tosi et al.~\cite{NLA} use a confidence map in order to detect reliable and unreliable points in a disparity map. After removing the unreliable points from the map, they then update these unreliable points by aggregating the first reliable points along different paths that they call "anchor". They weigh each anchor according to a Gaussian similarity function and finally take a weighted median to update the unreliable point.

S. Gidaris and N. Komodakis~\cite{reg:drr} use three steps in order to improve the final disparity prediction. First they \textbf{detect} erroneous disparities by taking the initial estimate and performing a consistency check. Then they \textbf{replace} these inconsistently labeled pixels with a new label which is produced by a convex combination of the initial label field. In the end they \textbf{refine} the disparity map by doing a residual correction in order to get a ''softer'' output with finer structures.

Additionally to cost-aggregation/belief propagation, disparity refinement often includes subpixel refinement~\cite{reg:subpx} a consistency check~\cite{reg:subpx_and_cons}\cite{reg:cons_check1}\cite{reg:cons_check2} and hole-filling/gap interpolation~\cite{reg:int_gaps}.

Our work differs from prior work on stereo vision in (1) the network structure for feature extraction and (2) the post-processing step. Prior work often relies on deep structures with many trainable weights and out-of-the-box disparity refinements like semi-global matching or conditional random fields. We address these issues by using a dense network structure with a three-step disparity refinement procedure.

\section{Network}

\begin{table}
\center
\caption{Layer ablation study}
\label{tab:layer_ablation}
\begin{tabular}{|c|c|c|}
\hline
layers & parameters & 2-PE\\ 
\hline
2-layer & 37k & 49.08\\
\hline
3-layer & 111k & 41.23\\
\hline
4-layer & 222k & 37.32\\ 
\hline
5-layer & 369k & 35.90 \\ 
\hline
6-layer & 554k & 35.15\\ 
\hline
\end{tabular}
\end{table}
In this section, we describe the network architecture of our model.
We use fully-convolutional siamese branches with shared weights for our network. This network consists of five convolutional layers with a kernel size of $k = 3x3$ and $m = 64$ feature maps per layer. We took inspiration from DenseNet by G. Huang et al.~\cite{densenet} by connecting the output of each layer to each subsequent layer. While this network was originally build with the task of object-detection in mind, we argue that this structure is a good fit for disparity estimation as well.
The benefits described in their work, such as strengthened feature propagation and better feature reuse while alleviating the vanishing gradient problem leads to a more lightweight feature extractor. However we adapt the original implementation to fit our needs. The changes to the original structure are as follows:

\begin{itemize}
    \item Following Zbontar and LeCun~\cite{disp:mc_cnn} we do not use down-sampling in our network. Therefore we do not use ''Dense Blocks'' and transition layers as described by G. Huang et al.~\cite{densenet}. The transition layer is described as having a $1x1$ convolution as well as down-sampling and batch-normalization. All layers in between these transition layers are called a ''Dense Block''.
    \item In the original work of G. Huang et al.~\cite{densenet} the structure is described as being deep and narrow, e.g. having only $12$ filters per layer but having many layers. We decided on a shallower and wider network structure, with $64$ filters per layer. The amount of input connections increases linearly with $2*m$ input weights in the third layer and up to $4*m$ in the output layer.
    \item The original implementation uses ReLU~\cite{act:relu} as their activation function. Related work~\cite{disp:tanh1}\cite{disp:tanh2}\cite{disp:cnn_crf} has shown that using a TanH activation function gives better results than using ReLUs~\cite{act:relu} for feature matching. Therefore we also use TanH as our activation function.
\end{itemize}

\begin{table}[t]
\renewcommand{\arraystretch}{1.3}
\caption{Number of trainable parameter comparison of popular networks}
\label{tab:param}
\centering
\begin{tabular}{|c|c|}
\hline
Method & Param\\
\hline
FC-DCNN (ours) & 0.37M \\
\hline
MC-CNN-ACRT~\cite{disp:mc_cnn} & 0.5M\\
\hline
GC-Net~\cite{disp:gc_net} & 2.9M \\
\hline
PSMNet~\cite{disp:psm_net} & 3.5M \\
\hline
\end{tabular}
\end{table}

The total number of parameters for the feature extractor network is around 370k in 5-layers. To motivate this architecture, we conduct an ablation study by increasing the number of layers from 2 to 6. We train each network overnight and report the number of trainable parameter as well as the end-point error with a threshold of two (2-point error or 2-PE) on the Middlebury dataset. The results of this study can be seen in Tab.~\ref{tab:layer_ablation}. This experiment shows that the network accuracy improves noticeably by increasing the number of layers up until 5 layers. After that point adding another layer seems to affect the overall accuracy less, however the number of trainable parameters increases drastically. We therefore decided on a 5-layer network structure as illustrated in Fig.~\ref{fig:network_arch}.
Tab.~\ref{tab:param} compares the number of parameters of our network (FC-DCNN) to some other popular methods. 
This illustrates that our network is a very lightweight network and therefore easily extendable, but still produces comparable results on challenging outdoor and indoor datasets.

\subsection{Post-processing}
In order to produce the final result, some post-processing is necessary.
The following steps are done in order to improve the final disparity estimation: First we filter each dimension of the cost-volume, then we find and remove inconsistent points. Last we replace this points with new values. This update changes depending on if the value is part of the foreground or background. We argue that post-processing is an important step of the method. Although it is responsible for most of the overall runtime, it further decreases the error by almost half. We motivate this by reporting on the 2-point error and runtime on the Middlebury training dataset for each post-processing step. The runtime and accuracy will always be given as all the previous steps in addition to the step currently described. Without any post-processing the network achieves a 2-point error of $33.3\%$ with an average execution time of $3$ seconds.

\subsubsection{Filtering}
Median filter is known to work well for Salt\&Pepper noise~\cite{S_and_P}. This kind of noise is common in flat or textureless regions in disparity maps, even with learned, more expressive features. To get rid of some of this noise we apply a median filter with a $5x5$ kernel to each dimension of the cost-volume consecutively. To this end the differentiable and time-efficient implementation of the median-filter from the Kornia library is used~\cite{kornia}.

Afterwards each slice of the cost-volume is filtered by a guided filter with a radius size of $r= 8$ and a regularization parameter $\eta = 10$ to produce a final smooth output. We use a fast deep-FCN implementation with pre-trained weights implemented in pytorch for this task~\cite{guided_filter}. This decreases the error on Middlebury from $33.3\%$ to $22.6\%$ with an average execution time of $12.1$ seconds. 

\subsubsection{LR-consistency check}
The left-right consistency aims to get rid of inconsistencies between the disparity map calculated for the left image and the disparity map calculated for the right image.
These inconsistencies are expected to occur because of self-occlusions, for instance at the object-boundaries, however they also occur if the match is predicted wrong.

The check is done by also treating the right image as reference and searching corresponding image points in the left image along the opposite search direction. 

Let $D^{L}$ be the disparity map obtained by treating the left image as reference and $D^{R}$ the disparity map obtained from treating the right image as reference. Let furthermore $d$ be the value of $D^{L}$ at position $(x,y)$, i.e. $d = D^{L}(x,y)$ Then a value is marked as inconsistent if:

\begin{equation}
   |D^{L}(x,y) - D^{R}(x-d,y)| > 1.1.
\end{equation}

This step doubles the execution time, as all of the steps have to be done for the left and for the right image individually (plus the runtime for the consistency check itself). On Middlebury this increases the runtime from $12.1$ seconds to $21.6$ seconds. This step does not improve the accuracy, however it produces a disparity map with removed inconsistencies.
\subsubsection{Update inconsistent points}
Once all inconsistent points in a disparity map have been found, their values should be updated. We argue that inconsistencies occur because of either: 1) self-occlusion of an object, 2) that the structure is outside of the field-of-view of the second image, in which case you cannot find the right prediction with the data alone, or 3) simply put the prediction was wrong.

We further argue that image points that are part of the background are most likely flat and because of the epipolar constraint will therefore have most likely the same depth as points in the same horizontal line. If we find a pixel marked as invalid that is part of the background, we search for the first valid measurement on the same horizontal line that is also part of the background and copy it. If the end of the horizontal line is reached, the direction is reversed and a valid background measurement is searched to the left. 

However, the same cannot be said about invalid pixels that are part of the foreground, as it contains complex structured objects. If an invalid point is part of the foreground, an averaging approach is taken. Here we search in all eight cardinal directions, starting from the invalid point until a valid point, that is also part of the foreground, is found along this scanline. Afterwards all eight values are summed up and averaged for the new disparity value of this point.

\begin{figure}[!t]
\centering
\includegraphics[width=2cm]{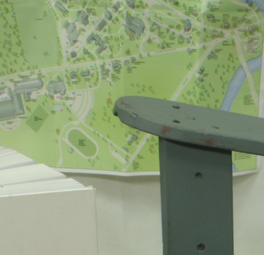}
\includegraphics[width=2cm]{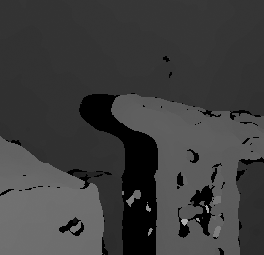}
\includegraphics[width=2cm]{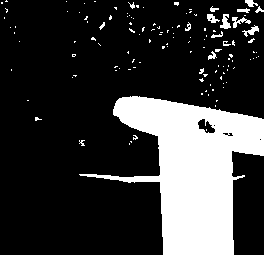}
\includegraphics[width=2cm]{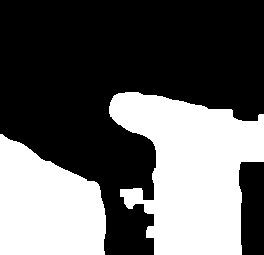}

\caption{From left to right: RGB image of an image detail, disparity map with inconsistencies removed (black), watershed mask on RGB image, watershed mask on disparity map (white = foreground, black = background).}
\label{fig:segmentation}
\end{figure}

In order to get the foreground and background segmentation, a simple watershed algorithm is used on the disparity with removed inconsistencies. This is because we want image points that are outside the field-of-view of the right image to be treated as background pixels, as averaging would not make sense in this case. Furthermore, object boundaries of the disparity image are often not exactly were they are in the RGB image. An example can be seen in Fig.~\ref{fig:segmentation}. This small area was chosen to illustrate that the overlap of the object boundary between the RGB image and the corresponding estimated disparity map is not perfect and therefore the mask obtained from the RGB image (second to last image in Fig.~\ref{fig:segmentation}) would wrongly classify disparities as part of the foreground. Therefore we use the disparity map with deleted inconsistent points as an input (second image in Fig.~\ref{fig:segmentation}) to produce the final mask (last image in Fig.~\ref{fig:segmentation}). This means however, that larger holes within foreground structures will be classified as background. To close such holes in the mask we use a dilation scheme with a $5x5$ filter kernel size for two iterations. Afterwards we thin the mask again with an erosion scheme for two iterations with a $5x5$ kernel.

On Middlebury this step increases the accuracy from $22.6\%$ to $17.9\%$ with an average execution time of $27.6$ seconds.

Depending on the dataset, this method is 4 to 5 times faster than MC-CNN-acrt~\cite{disp:mc_cnn} improving the runtime from $106$ to $27$ seconds on the Middlebury dataset. This time measurements have been taken directly from the corresponding official benchmarks.

\section{Experiments and results}
In this section we discuss all the conducted experiments and their results. It is structured as follows: First we will discuss how the training task is defined and subsequently how the training data is prepared for this task. Afterwards we will discuss the implementation details. Finally we will show the qualitative and quantitative results of our experiments and compare them to other methods.

\subsection{Training the feature extractor}
Disparity estimation can be viewed as a multi-classification problem. For each position in the reference image there is a (previously fixed) number of candidates corresponding to a possible position in the second image. In the end the most likely candidate is chosen (winner-takes-all) for the final prediction. However, instead of directly predicting the final winner, for instance by calculating the Cross Entropy loss over all possible classes, a simpler approach is taken in order to train the feature extracting siamese network. 

Following Zbontar and Lecun's work~\cite{disp:mc_cnn}, we instead train the network as a binary classification task. For each sample a small grayscale patch is extracted at position $p = (x,y)$ from the reference image. From the second image, two patches are extracted, one positive example $q_{pos}$ at the correct position and one negative example $q_{neg}$ at the wrong position.

\subsection{Preparing the training set}

As suggested by Zbontar and LeCun~\cite{disp:mc_cnn} we choose a patch size of $11x11$ for the randomly cropped patches of our training set. 
The center position of the left patch $p$ is randomly chosen over the whole image domain, as long as the corresponding ground truth $gt$ position is valid.
\begin{equation}
   p = (x, y)
\end{equation}
\begin{equation}
    gt(x,y) = valid.
\end{equation}

The positive patch $q_{pos}$ is created by using the correct disparity $d$ of position $(x,y)$
\begin{equation}
   q_{pos} = (x-d, y).
\end{equation}
In the original paper by Zbontar and LeCun~\cite{disp:mc_cnn} a small offset is added to the position of the positive patch because the post-processing worked better with it. As we use different post-processing steps we do not use any random offset for the positive patch. 

For the negative sample a random offset $o_{neg}$ within the range of either $(-6,-2)$ or $(2,6)$ is chosen. In our implementation the probability of either being shifted to the left or to the right is 50\%.
\begin{equation}
   q_{neg} = (x-d + o_{neg}, y).
\end{equation}

The reason why the random offset is limited and not chosen from all possible positions is because it is expected that points far away from the true position have a low similarity score anyway. Points closer to the positive position can be more ambiguous and should therefore be learned to be more robust.

In the original paper, Zbontar and LeCun~\cite{disp:mc_cnn} describe that about 20\% of their training set for the Middlebury benchmark consists of samples where the lighting condition or shutter exposure changed between the image pairs. In our experiments we found having 10\% of training samples with these variations works better, as it seems that it introduced more noise otherwise. We do not use any data set augmentation, however by using the "perfect" and "imperfect" rectification with the variations in lighting and exposure for the Middlebury training set, we still end up with around 40 million training samples, although many of them will be similar as we allow a position to be drawn multiple time.

\subsection{Implementation details}

The whole project is implemented using Python3 and pytorch 1.2.1~\cite{pytorch}. The loss function is implemented as a hinge-loss: \begin{equation}
    loss = max(0, 0.2 + s_{-} - s_{+}),
\end{equation}
where $s_{-}$ is the similarity score between the left patch $p$ and the patch from a wrong position of the right image $q_{neg}$. $s_{+}$ is the similarity score between $p$ and the patch from the correct position $q_{pos}$. This loss forces the network to train features, such that the similarity between correct matches should be higher by at least $0.2$.

In our implementation we slightly change the formulation of this loss by using a ReLU~\cite{act:relu} instead of the traditional hinge loss for ease of implementation. This leads to a sign change in the loss definition:

\begin{equation}
    loss = ReLU(s_{+} - s_{-} - 0.2).
\end{equation}

The similarity between patches is calculated using the pytorch implementation of the cosine similarity. It is defined as:
\begin{equation}
    sim(A,B) = \frac{A \cdot B}{\norm{A}\norm{B}},
\end{equation}
where $A$ and $B$ are the two vectors to be compared.
We us the Adam Optimizer~\cite{adam} with a relative small learning rate of $\eta = 6 \times 10^{-6}$ for training. We use a batch-size of $800$ samples per iteration and trained for roughly 2 days on each dataset using a GeForce RTX 2080.

\subsection{Results}

We compare our results to state-of-the-art methods from three challenging benchmarks.
For each benchmark the network was trained for around 2 days. We decided to train KITTI2012 and KITTI2015~\cite{kitti} together, however better results may be achieved by training and testing on each dataset individually.
\subsubsection{Middlebury}
\begin{table}
\renewcommand{\arraystretch}{1.3}
\caption{Accuracy comparison on the Middlebury training dataset}
\label{tab:results_mb}
\centering
\begin{tabular}{|c|c|c|c|c|c|}
\hline
Method & 4-PE & 2-PE & 1-PE & 0.5 PE \\ 
\hline
FC-DCNN (ours) & 12.3 & \textbf{17.9} & \textbf{34.7} & 65.1 \\
\hline
iResNet~\cite{iresnet} & \textbf{11.1} & 20.3 & 35.1 & 58.7 \\
\hline
SGM (Q)~\cite{reg:sgm} & 12.9 & 21.0  & 37.3 & 64.6 \\
\hline
PSMNet~\cite{psmnet} & 13.1 & 23.0 & 40.2 & 64.9 \\
\hline
SGBM1 (H)~\cite{opencv} & 17.6 & 23.3 & \textbf{36.5} & 57.8 \\
\hline
\end{tabular}
\end{table}
The Middlebury stereo dataset~\cite{mb} consists of challenging indoor scenes with large disparity ranges under different, controlled exposure and lighting technique. The dataset is provided in full (F), half (H) and quarter (Q) resolution. We chose to submit in half resolution due to hardware constraints.
Tab.~\ref{tab:results_mb} shows that our method is already better than popular stereo methods, such as SGM~\cite{reg:sgm} on quarter (Q) resolution, as well as SGM~\cite{reg:sgm} on full resolution (F) (not shown in Tab.~\ref{tab:results_mb}) or OpenCV's reimplementation of SGM which they call SGBM~\cite{opencv}. It further shows that we are on par with well-known deep-learning methods such as iResNet~\cite{iresnet} or PSMNet~\cite{psmnet}.
On average our method took about $13$ seconds per Megapixel for the Middlebury data.

\begin{figure}[!t]
\centering
\includegraphics[width=2.8cm]{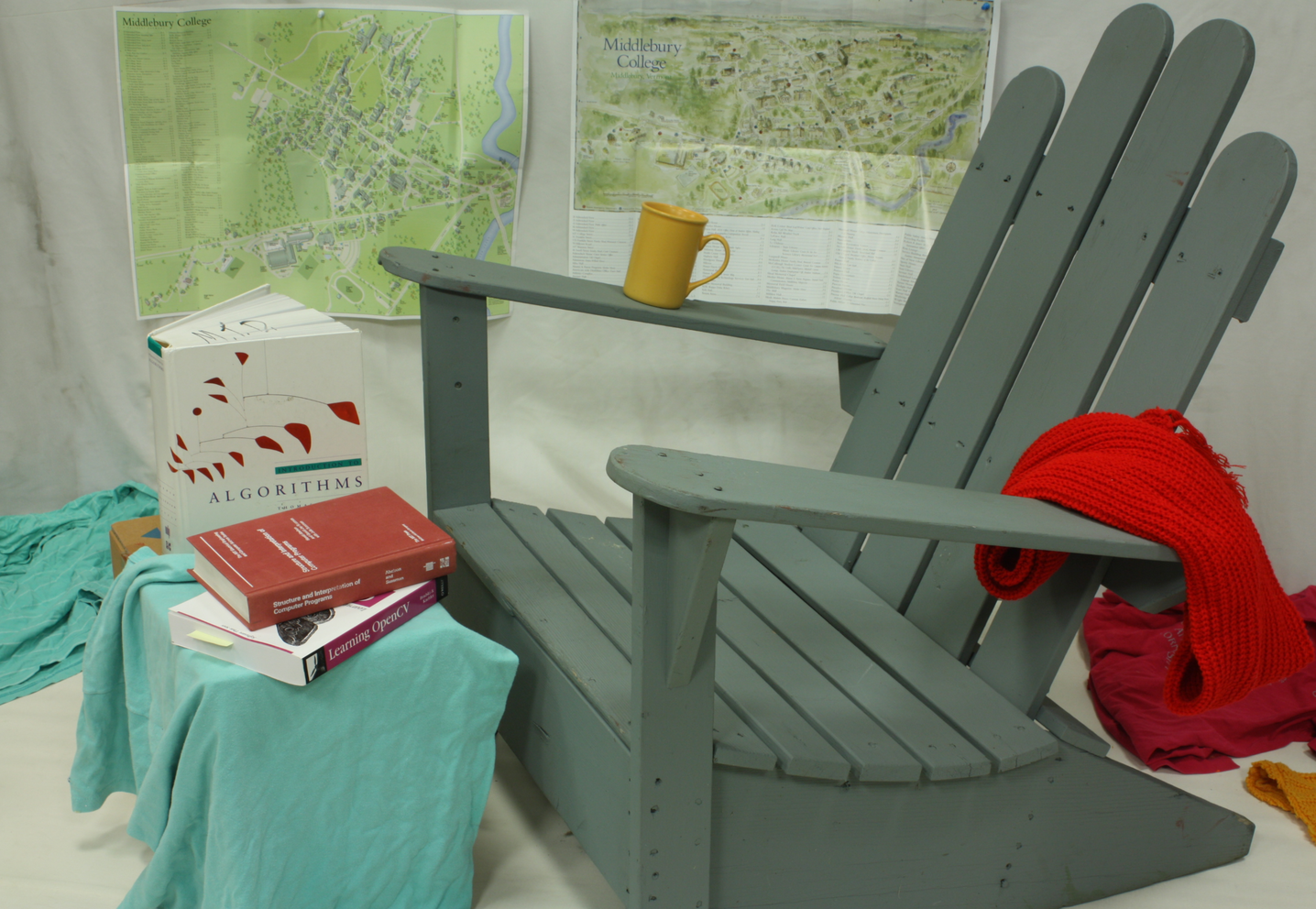}
\includegraphics[width=2.8cm]{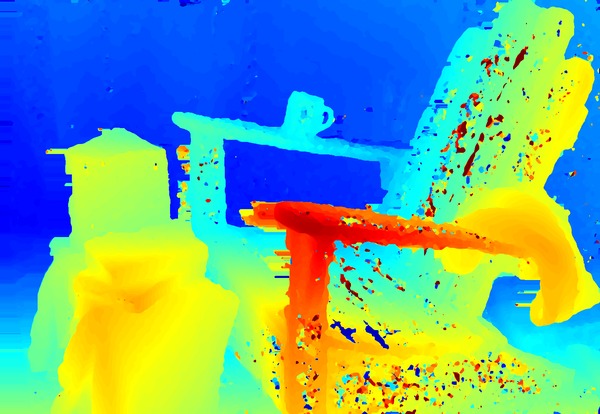}
\includegraphics[width=2.8cm]{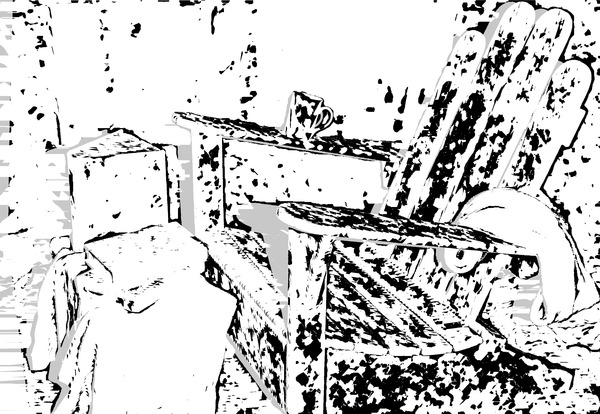}
\includegraphics[width=2.8cm]{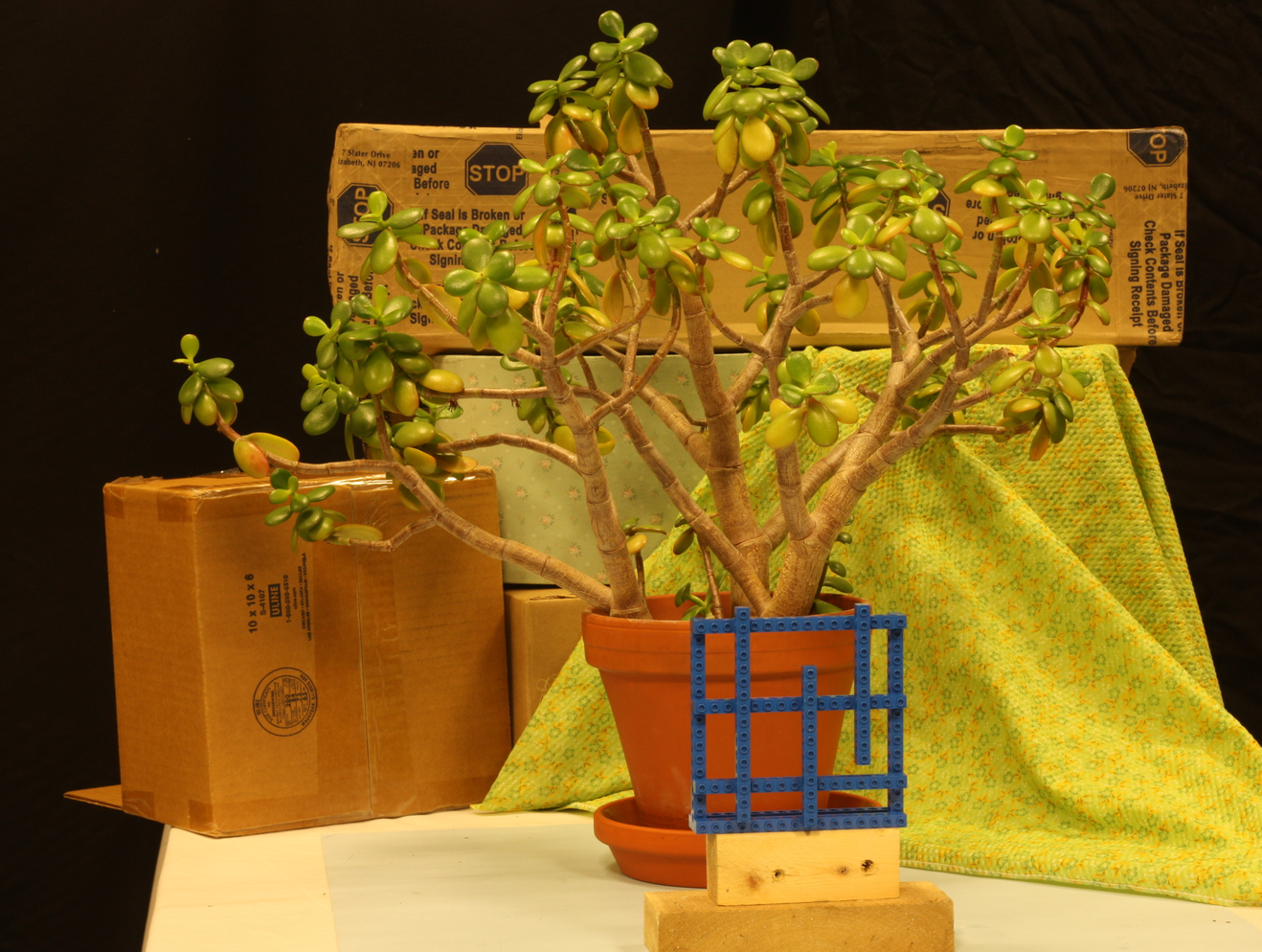}
\includegraphics[width=2.8cm]{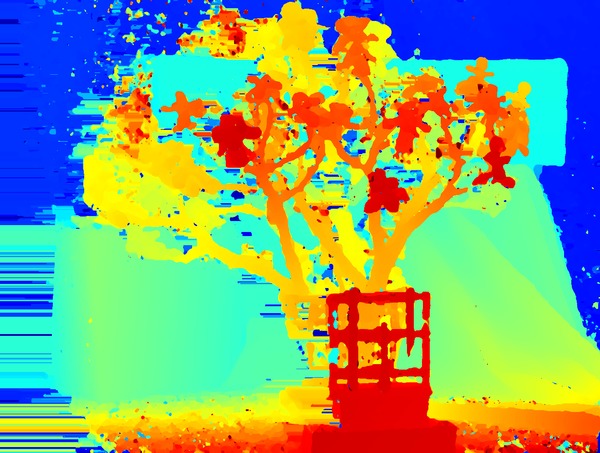}
\includegraphics[width=2.8cm]{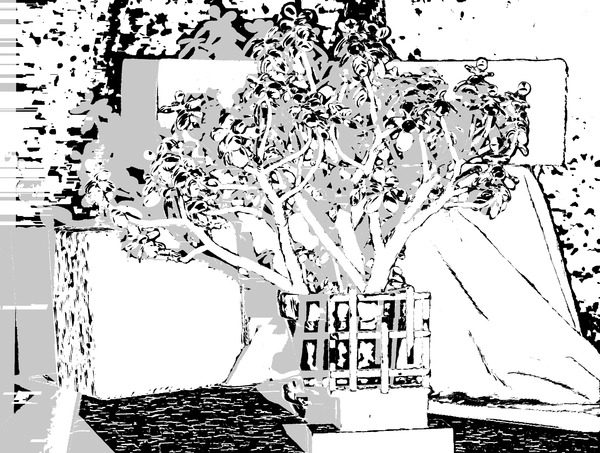}

\includegraphics[width=2.8cm]{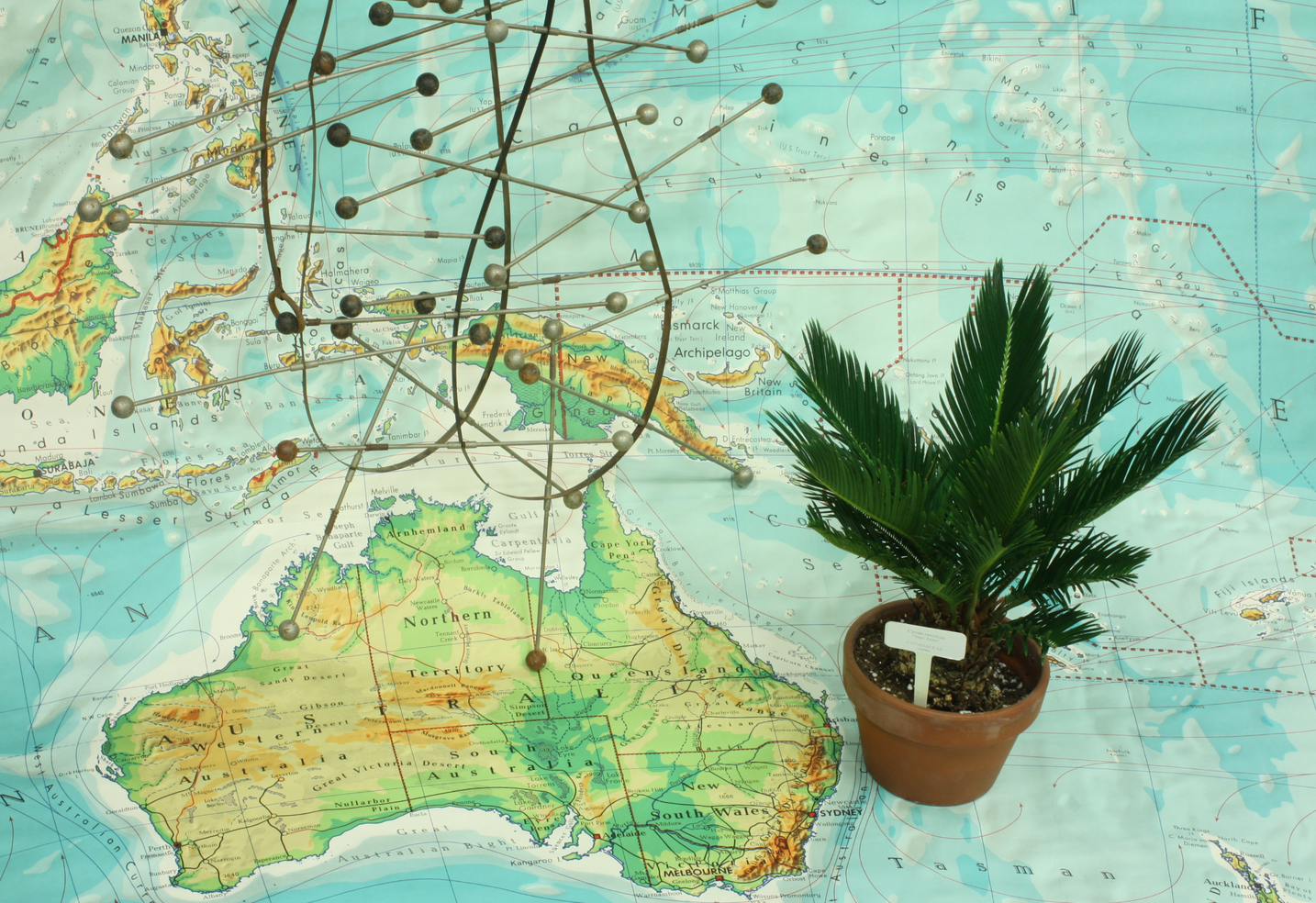}
\includegraphics[width=2.8cm]{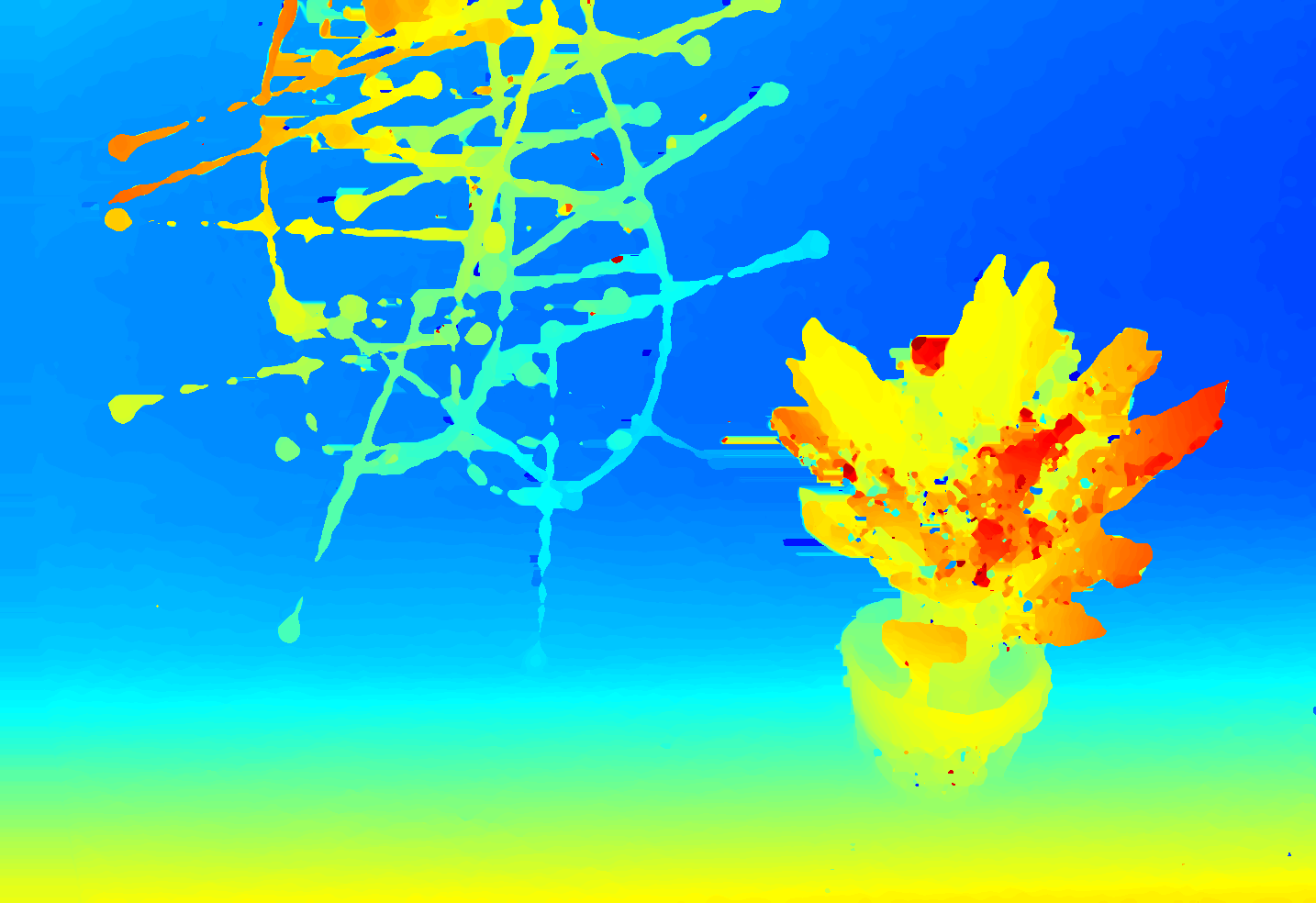}
\includegraphics[width=2.8cm]{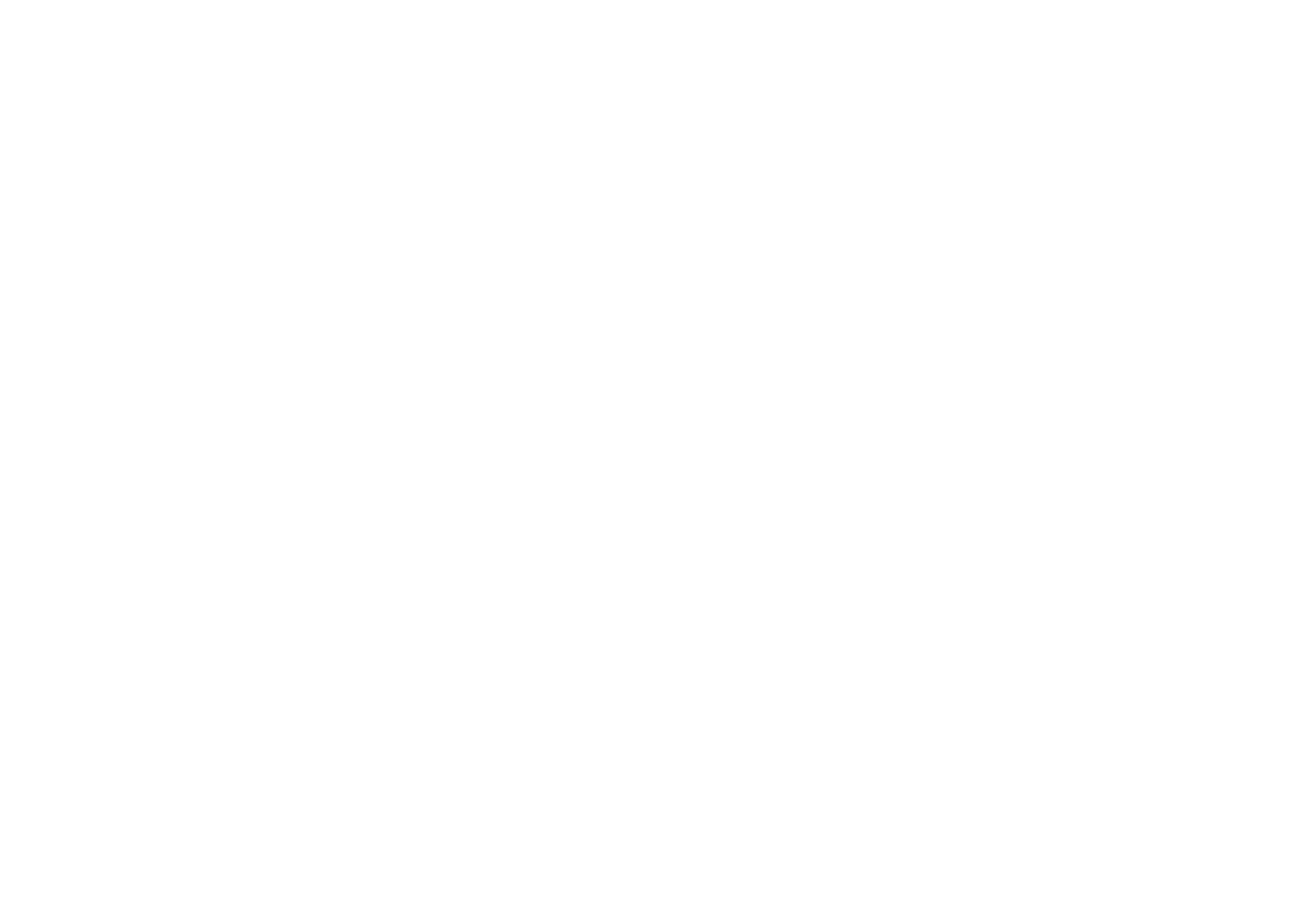}

\caption{Qualitative results from the Middlebury test and training dataset. From left to right: RGB, final disparity map, 2-point error on full resolution (not available for test data).}
\label{fig:error_mb}
\end{figure}

Fig.~\ref{fig:error_mb} shows qualitative results of our method of three different scenes from the Middlebury dataset. It illustrates that our method performs well even in cases of clutter, like the leaves of the ''Jadeplant'' sample (second example), or in homogeneous areas. An example for a homogenous area is the background area of the samples, here shown as blue colors in the disparity map. The last column consists of the 2-point error map obtained from the official submission page. Here darker colors correlate to a higher error.

\subsubsection{KITTI}
The KITTI stereo dataset~\cite{kitti} consists of outdoor street images used for autonomous driving. The ground truth was taken by a laser scanner which leads to a rather sparse ground-truth disparity. 
We use the same number of disparities for all pairs, namely $192$ for KITTI2012 and $228$ for KITTI2015.

\begin{table}[!h]
\renewcommand{\arraystretch}{1.3}
\caption{Accuracy comparison on the KITTI 2012 testing dataset}
\label{tab:results_kitti_test}
\centering
\begin{tabular}{|c|c|c|c|c|c|c|c|c|}
\hline
Method & 5-PE & 4-PE & 3-PE & 2-PE\\
\hline
FC-DCNN (ours) & \textbf{3.71} & \textbf{4.40} & \textbf{5.61} & \textbf{8.81}\\
\hline
OASM-Net~\cite{disp:oasm} & 4.32 & 5.11 & 6.39 & 9.01 \\
\hline
SGBM~\cite{opencv} & 5.03 & 6.03 & 7.64 & 10.60 \\
\hline
ADSM~\cite{disp:adsm} & 6.20 & 7.09 & 8.71 & 13.13 \\
\hline
GF (Census)~\cite{disp:gf_census} & 8.49 & 9.57 & 11.65 &  16.75\\
\hline
\end{tabular}
\end{table}

\begin{table}[!h]
\renewcommand{\arraystretch}{1.3}
\caption{Accuracy comparison on the KITTI 2015 testing stereo dataset}
\label{tab:results_kitti15_test}
\centering
\begin{tabular}{|c|c|}
\hline
Method & 3-PE\\
\hline
FC-DCNN (ours) & \textbf{7.71}\\
\hline
MeshStereo~\cite{disp:mesh_stereo} & 8.38 \\
\hline
OASM-Net~\cite{disp:oasm} & 8.98 \\
\hline
OCV-SGBM~\cite{sgm_mi} & 10.86  \\
\hline
SGM\&FlowFie+~\cite{sgm_fie} & 13.37 \\
\hline
\end{tabular}
\end{table}
Tab.~\ref{tab:results_kitti_test} compares our results on the KITTI 2012 test dataset with other methods. Tab.~\ref{tab:results_kitti15_test} compares our results on the KITTI 2015 test dataset~\cite{kitti} with other methods. Without any bells and whistles our method is better than well-known methods such as OpenCVs~\cite{opencv} implementation of SGM called SGBM or an implementation of Census features with guided filtering~\cite{disp:gf_census} while being on par with other state-of-the-art deep learning methods such as OASM-Net~\cite{disp:oasm}. We believe that doing further evaluations on the post-processing parameters would improve the results. On average the whole method takes about 7 seconds per image pair for the KITTI dataset~\cite{kitti}.

\begin{figure}[!t]
\centering
\includegraphics[width=3.9cm]{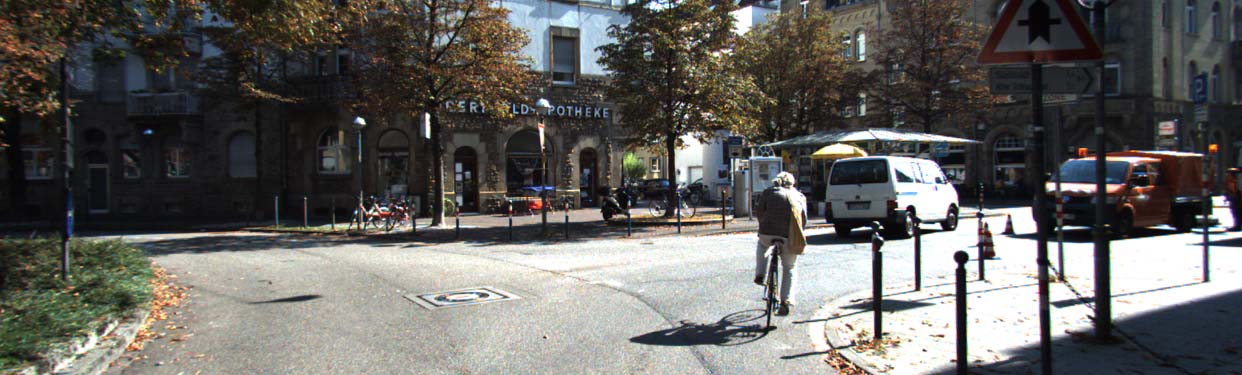}
\includegraphics[width=3.9cm]{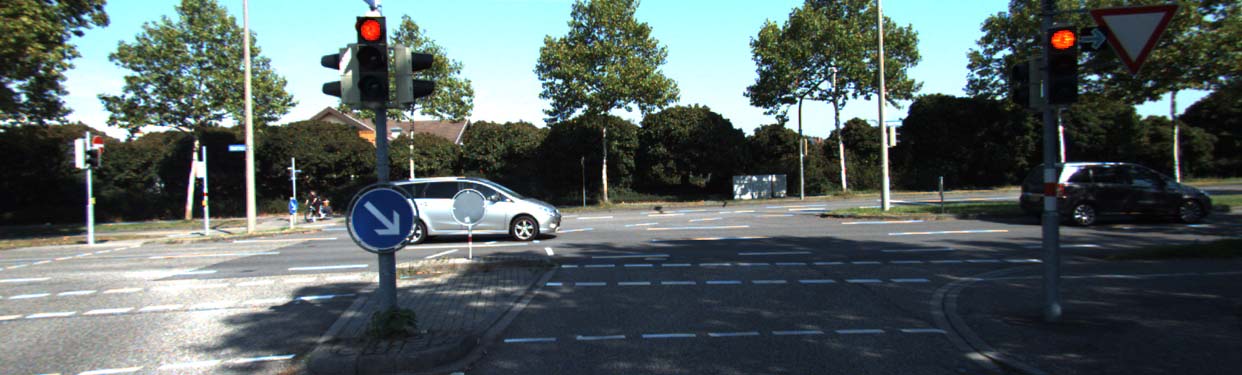}

\includegraphics[width=3.9cm]{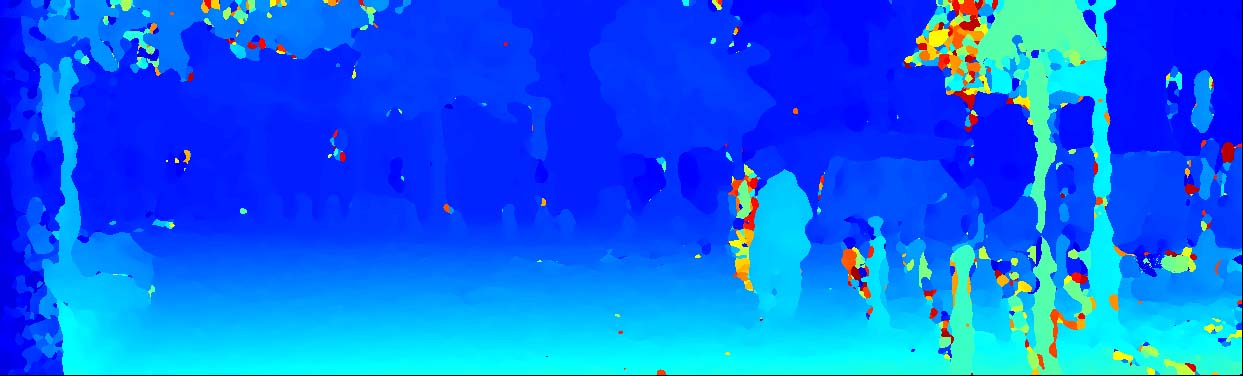}
\includegraphics[width=3.9cm]{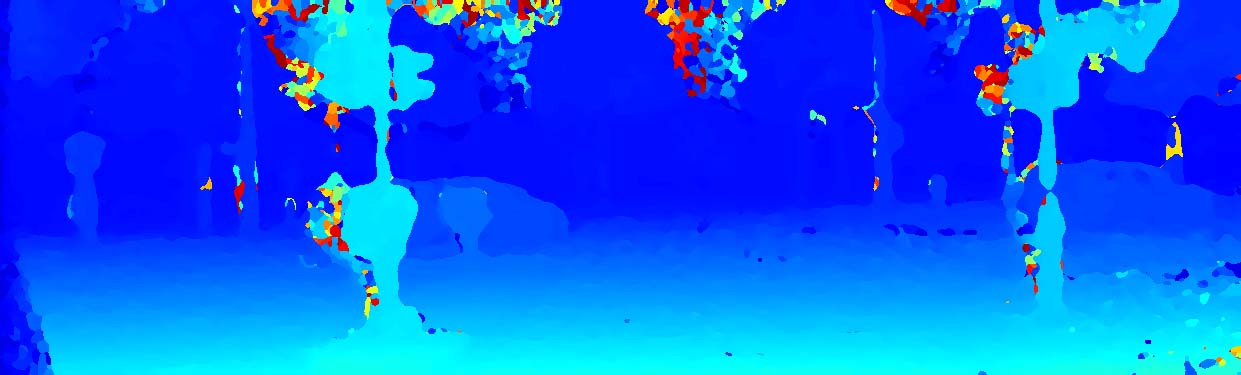}

\includegraphics[width=3.9cm]{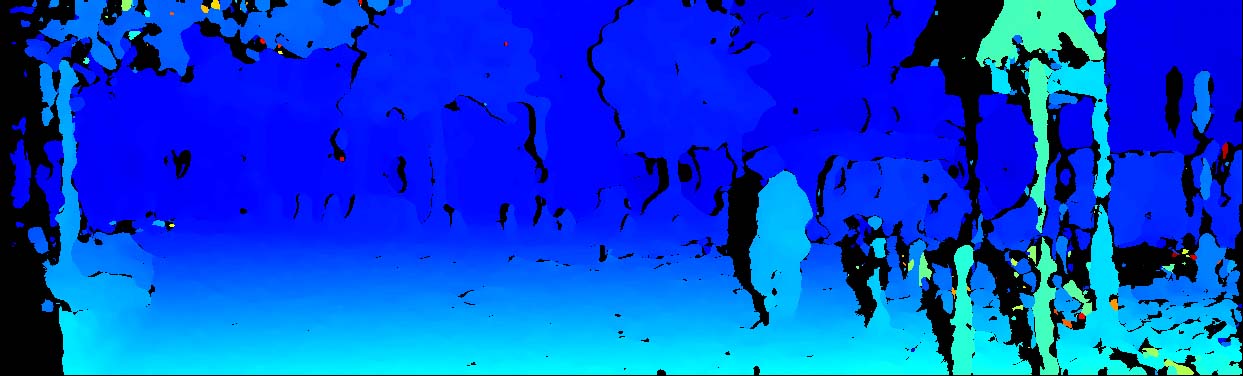}
\includegraphics[width=3.9cm]{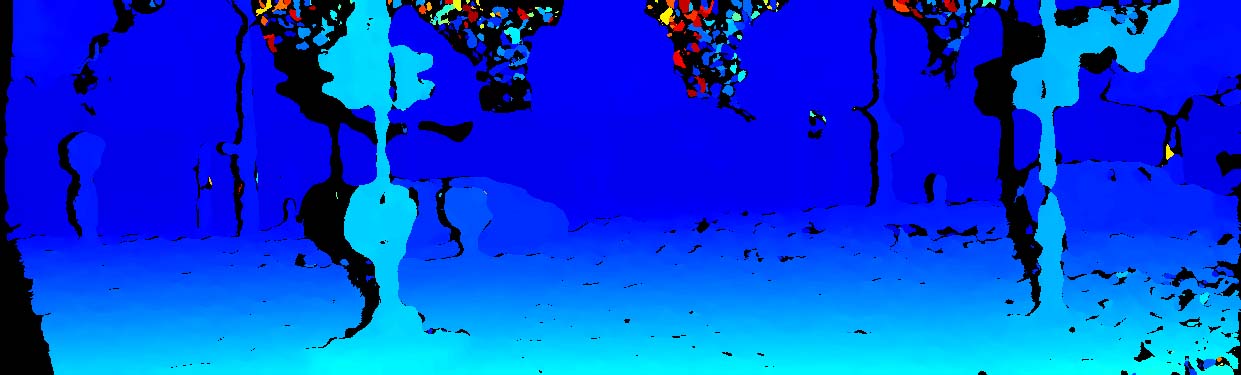}
\vspace{0.1cm}
\includegraphics[width=3.9cm]{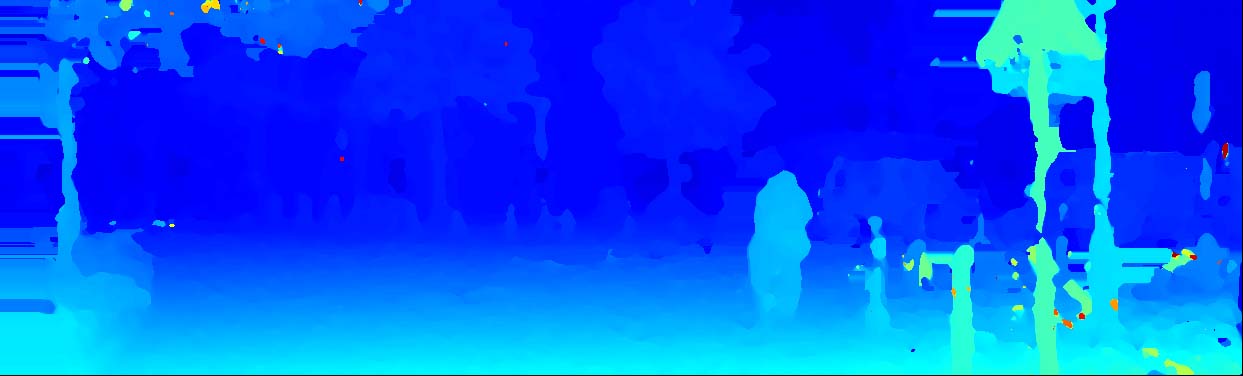}
\includegraphics[width=3.9cm]{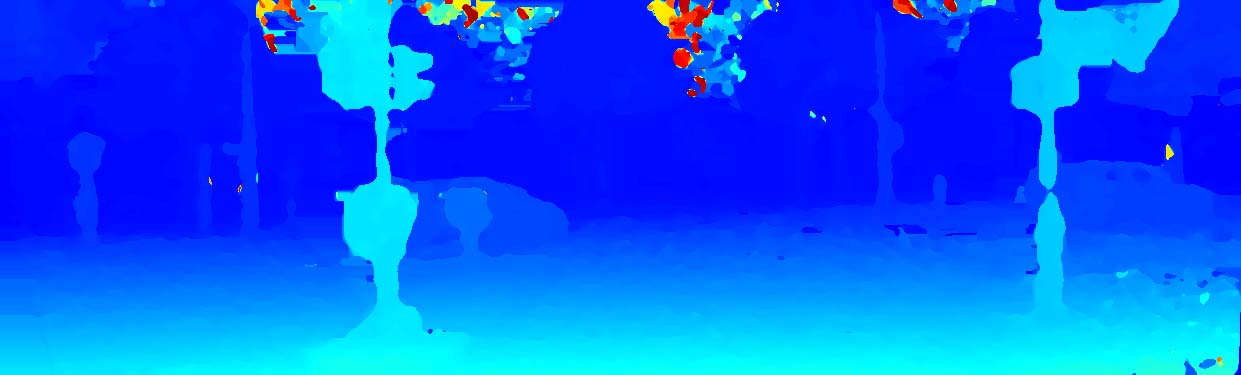}
\includegraphics[width=3.9cm]{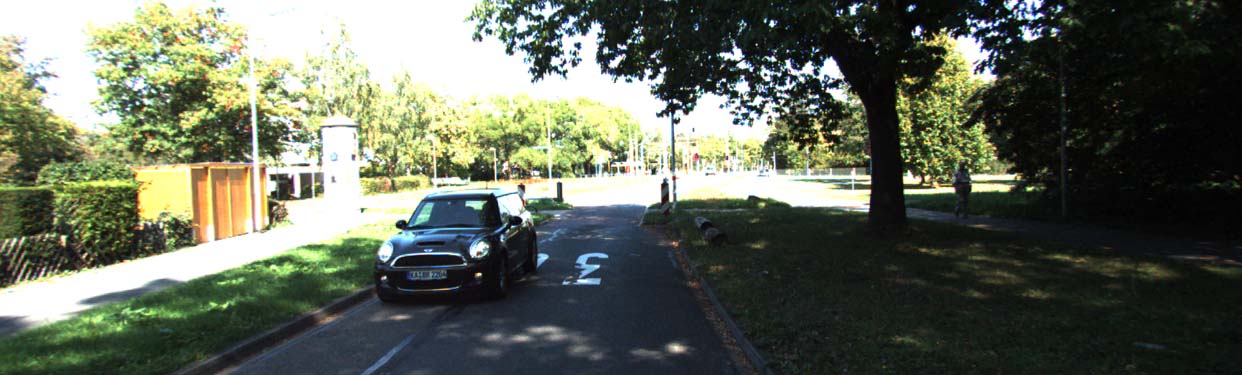}
\includegraphics[width=3.9cm]{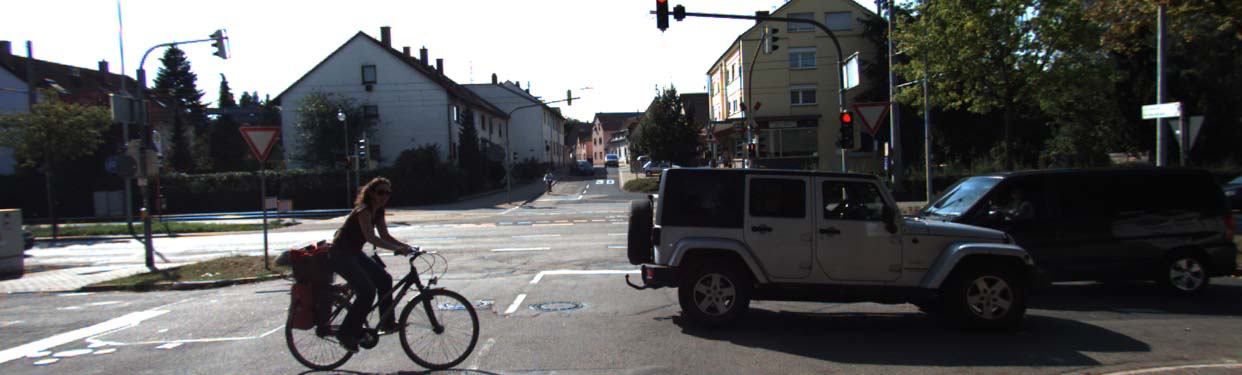}

\includegraphics[width=3.9cm]{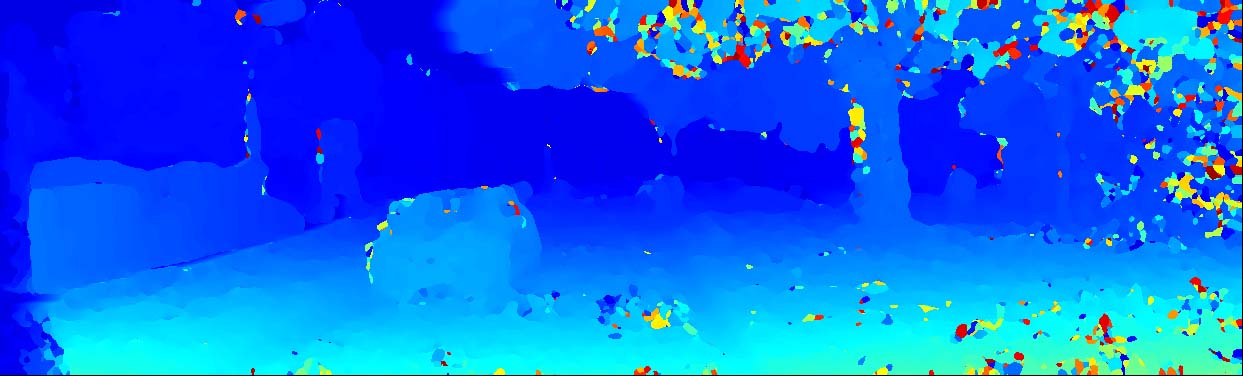}
\includegraphics[width=3.9cm]{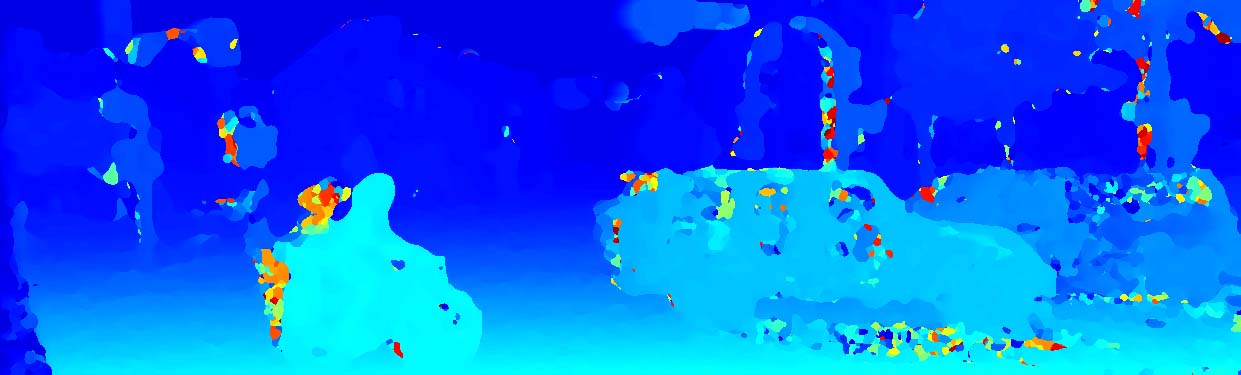}

\includegraphics[width=3.9cm]{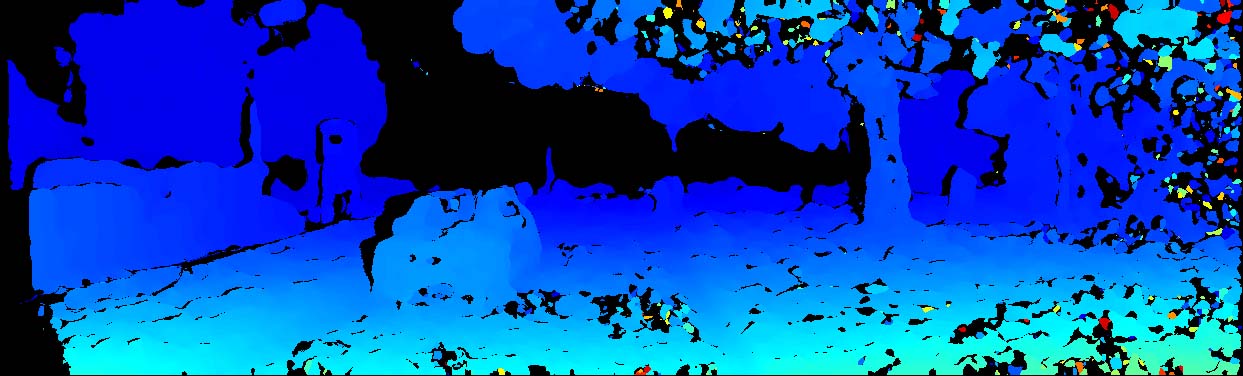}
\includegraphics[width=3.9cm]{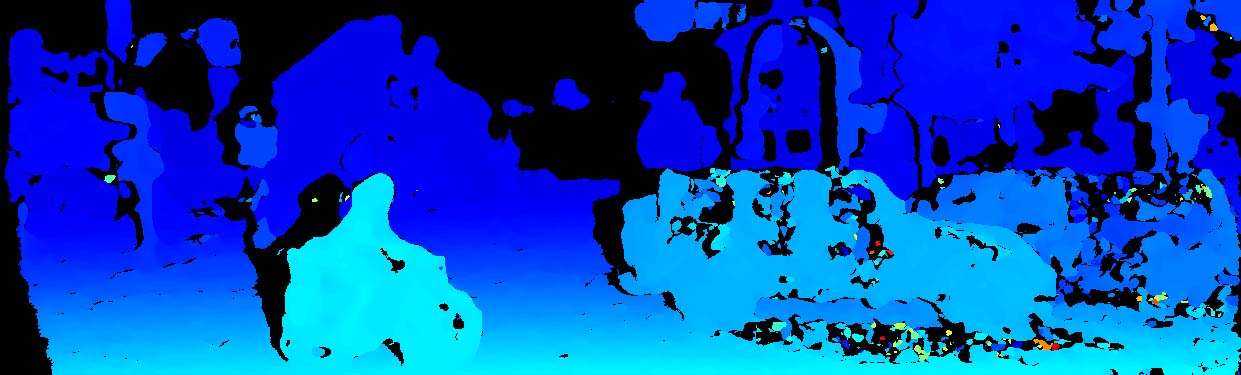}

\includegraphics[width=3.9cm]{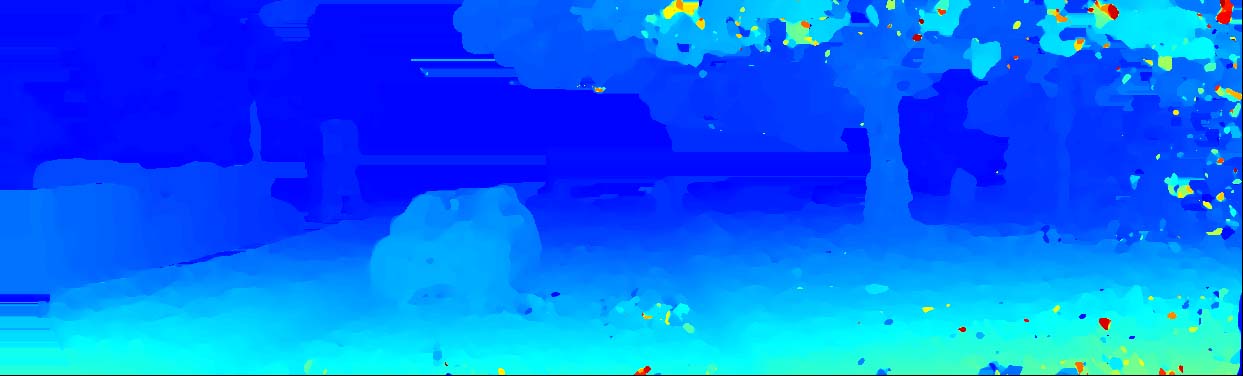}
\includegraphics[width=3.9cm]{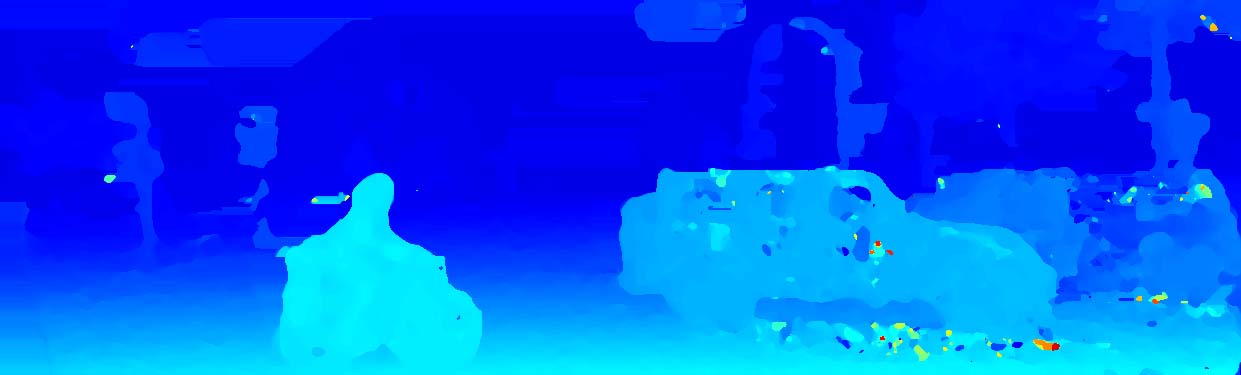}

\caption{Qualitative results from the KITTI train dataset (left column) and test dataset (right column). From top to bottom in both columns: left RGB image, initial disparity estimation, disparity with inconsistent points removed, final disparity map.}
\label{fig:results_kitti}
\end{figure}

Fig.~\ref{fig:results_kitti} shows the final disparity prediction plus all intermediate outputs of four chosen examples from the KITTI 2015 set~\cite{kitti}. It shows that while further hyperparameter studies might improve the accuracy, our method yields good results.

\subsubsection{ETH3D}
The ETH3D~\cite{eth3d} dataset for two-view stereo estimation consists of 27 training and 20 test pairs. The scenes of these pairs vary from tunnels to playgrounds and forest areas. In contrast to the other benchmarks however, the image-pairs have a small baseline and fine structures, which leads to a low number of disparities (maximum disparity in the training set is 64) with more steps in-between each discrete disparity step. We predict discrete-valued disparity steps, and also prepare our training patches as such. This would suggest, that our method is not well suited for this benchmark, however we show that we get decent results even with the previously stated limitations.

\begin{table}[!h]
\renewcommand{\arraystretch}{1.3}
\caption{Accuracy comparison on the ETH3D test dataset}
\label{tab:results_eth3d_test}
\centering
\begin{tabular}{|c|c|c|c|c|c|}
\hline
Method & 4-PE & 2-PE & 1-PE & 0.5 PE \\ 
\hline
FC-DCNN (ours) & 3.38 & \textbf{5.77} & \textbf{10.41} & 24.12 \\
\hline
MeshStereo~\cite{disp:mesh_stereo} & \textbf{2.61} & 5.78 & 11.52 & \textbf{22.27} \\ 
\hline
LSM & 4.58 & 7.38 & 14.01 & 29.98 \\
\hline
ELAS\_RVC~\cite{disp:elas} & 2.84 & 7.69 & 16.54 & 33.79 \\
\hline
\end{tabular}
\end{table}

Tab.~\ref{tab:results_eth3d_test} compares our method with other methods from the online leaderboard. The LSM methods is an anonymous submissions, therefore we cannot credit them. One can see that while there is still room for improvement, especially in the lower threshold end-point errors, our method already performs decently without any need of modification with only around 6\% 2-point error. On average our method took about $1.6$ seconds for one image pair to produce the final output.
\begin{figure}[!t]
\centering

\includegraphics[width=3.5cm]{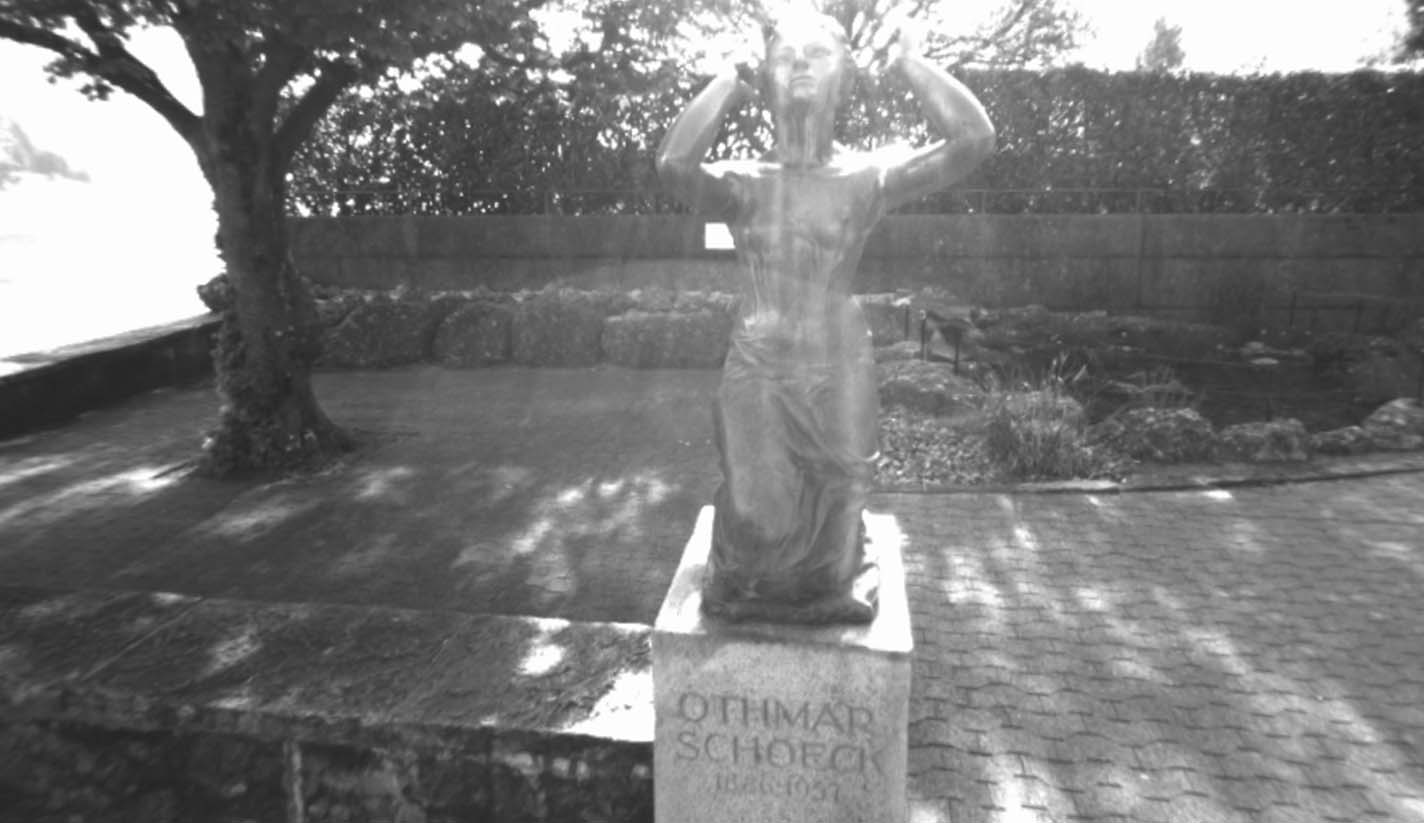}
\includegraphics[width=3.5cm]{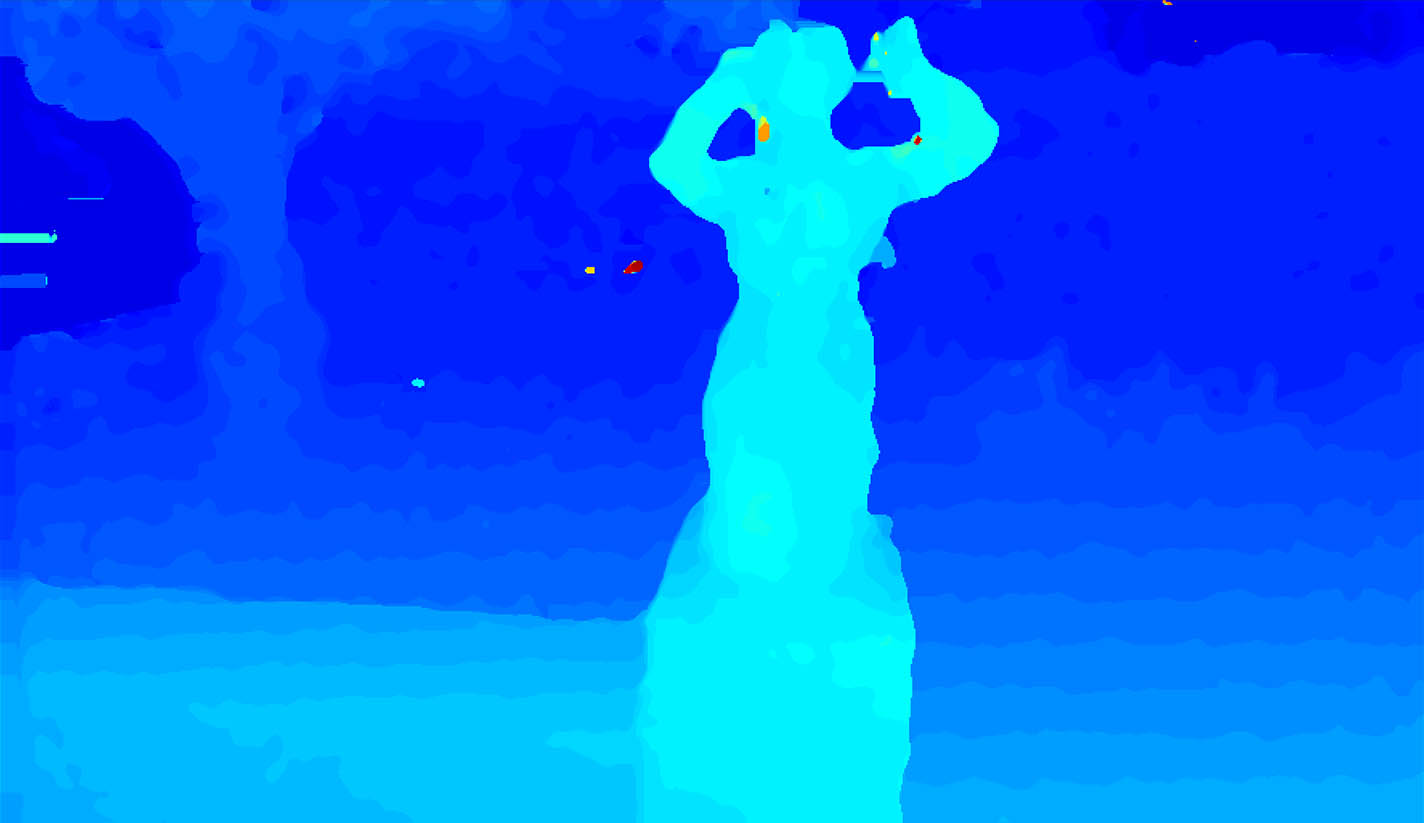}

\includegraphics[width=3.5cm]{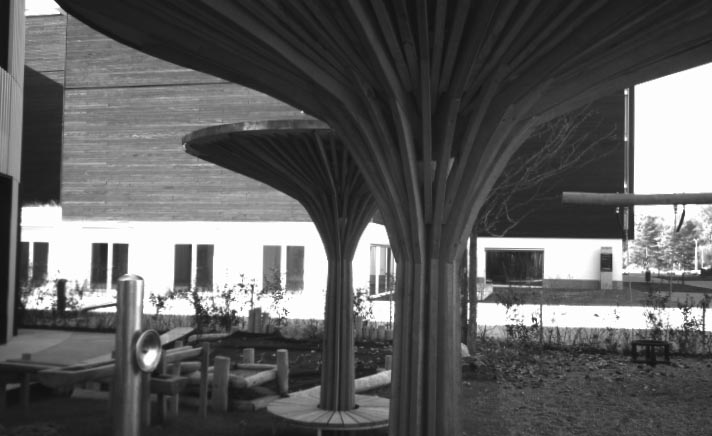}
\includegraphics[width=3.5cm]{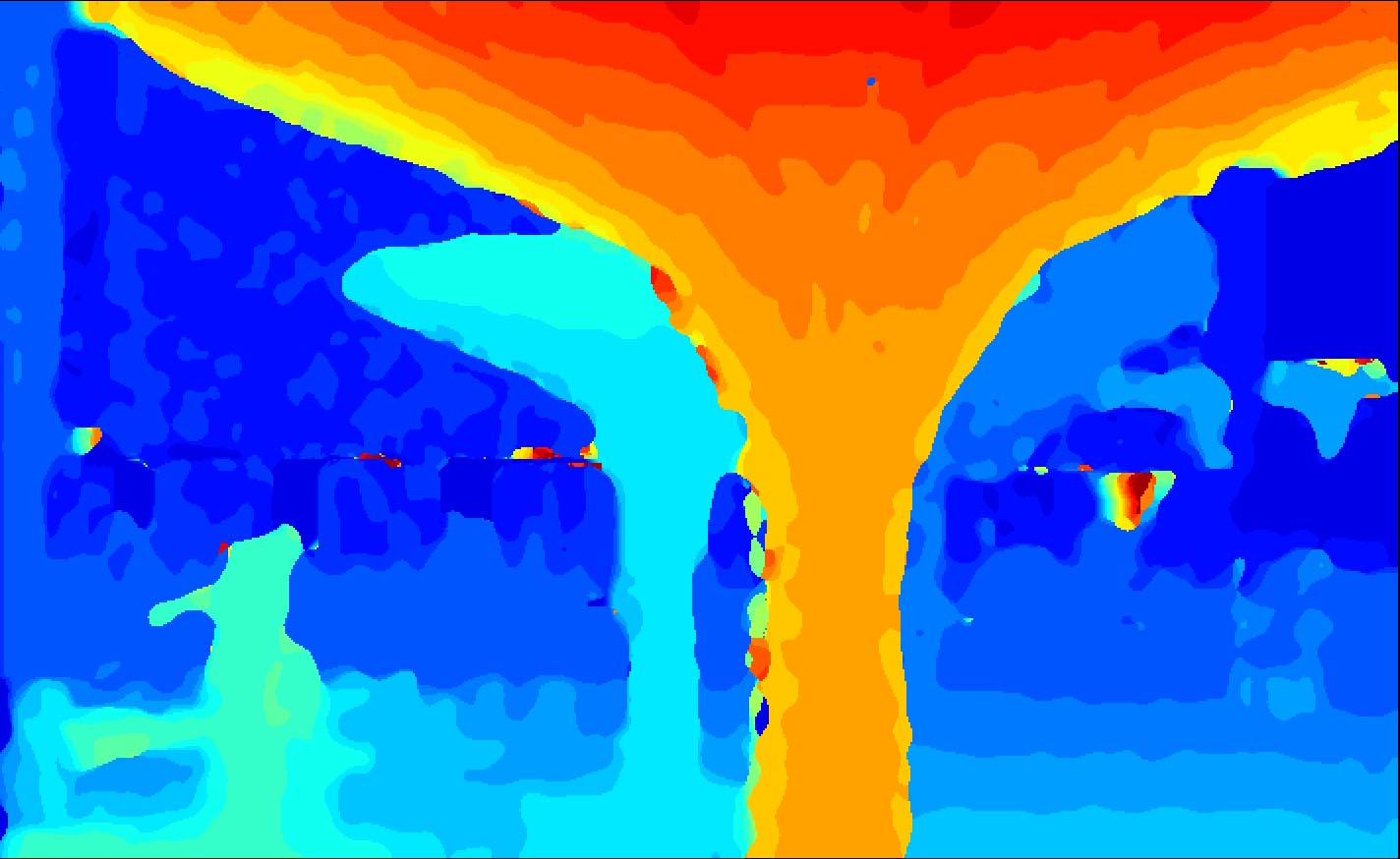}

\includegraphics[width=3.5cm]{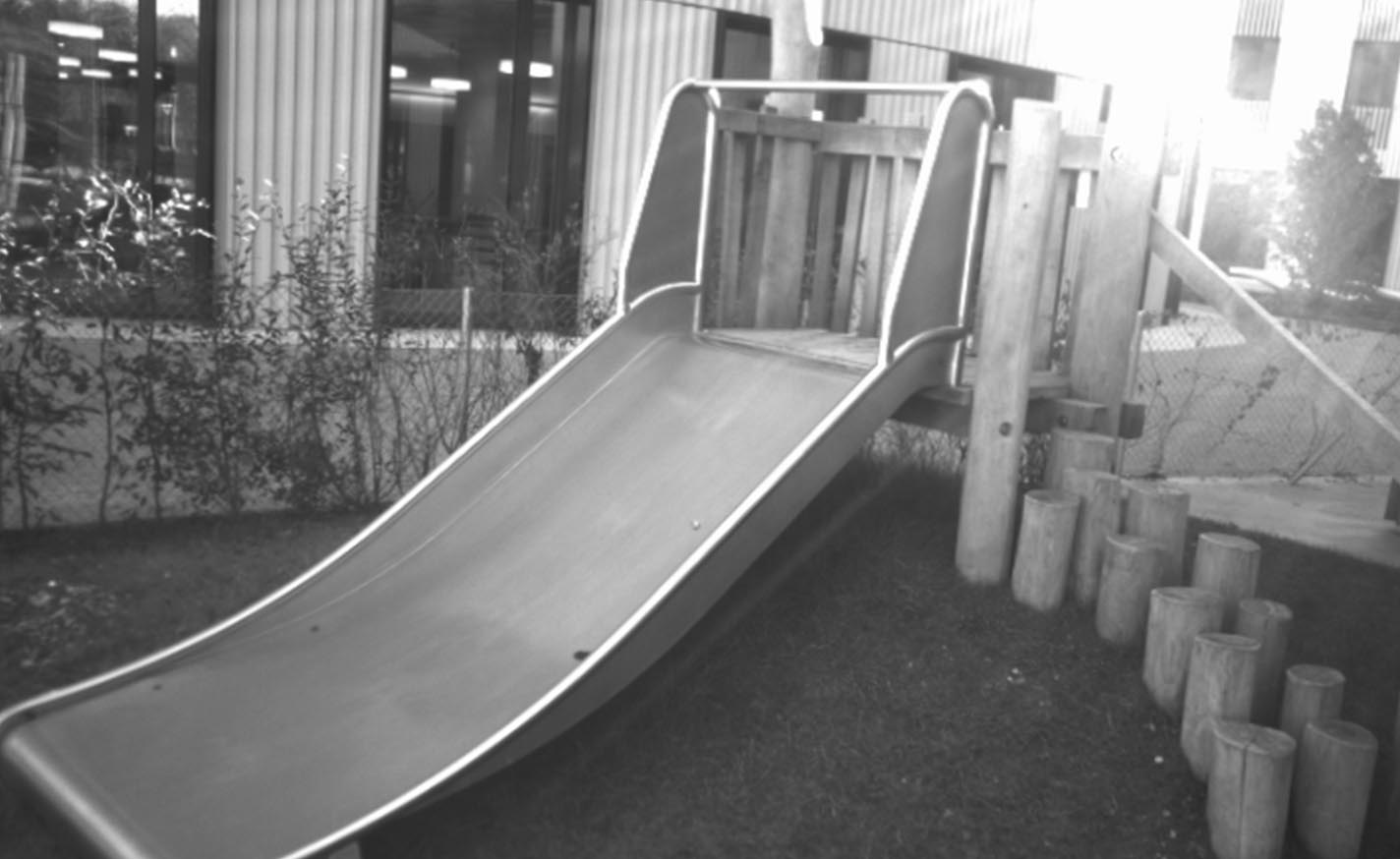}
\includegraphics[width=3.5cm]{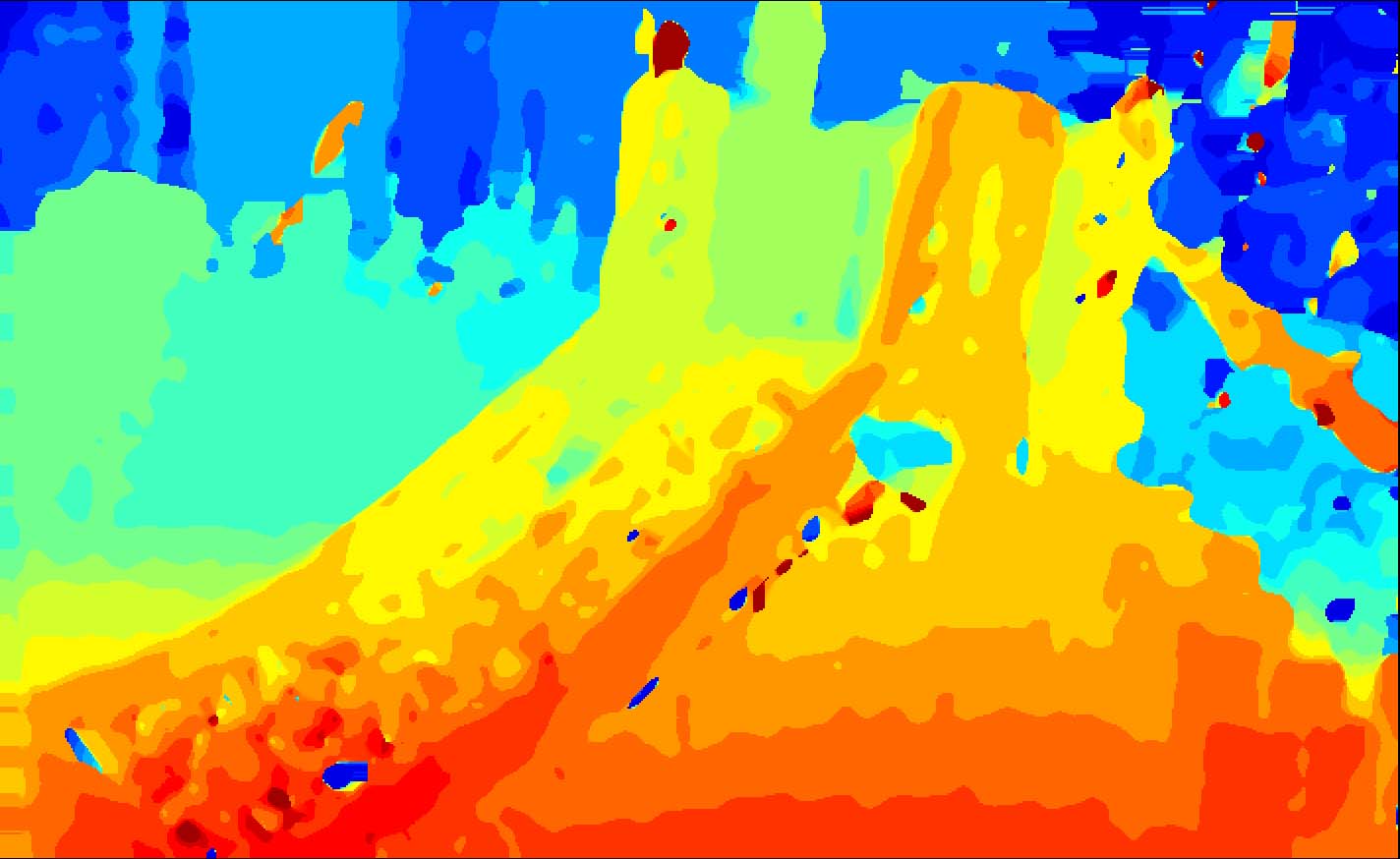}

\includegraphics[width=3.5cm]{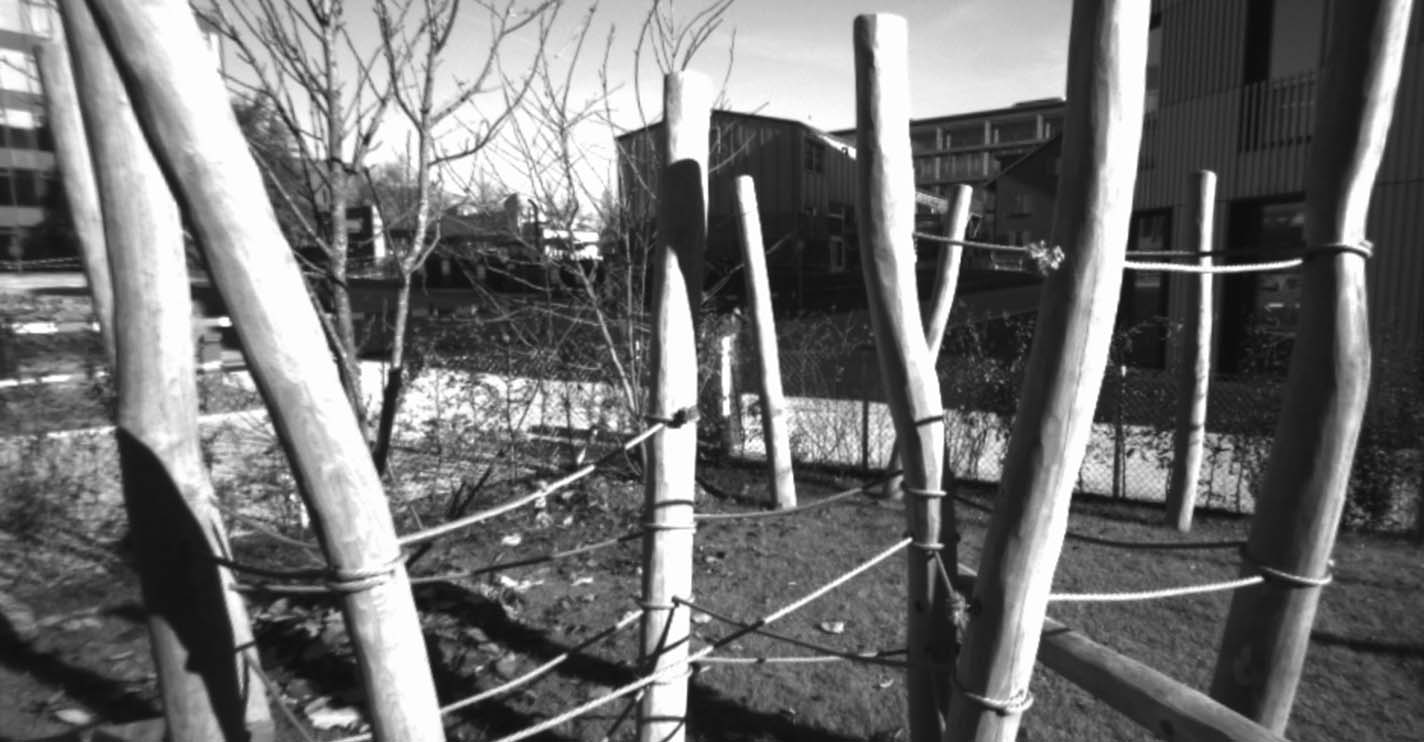}
\includegraphics[width=3.5cm]{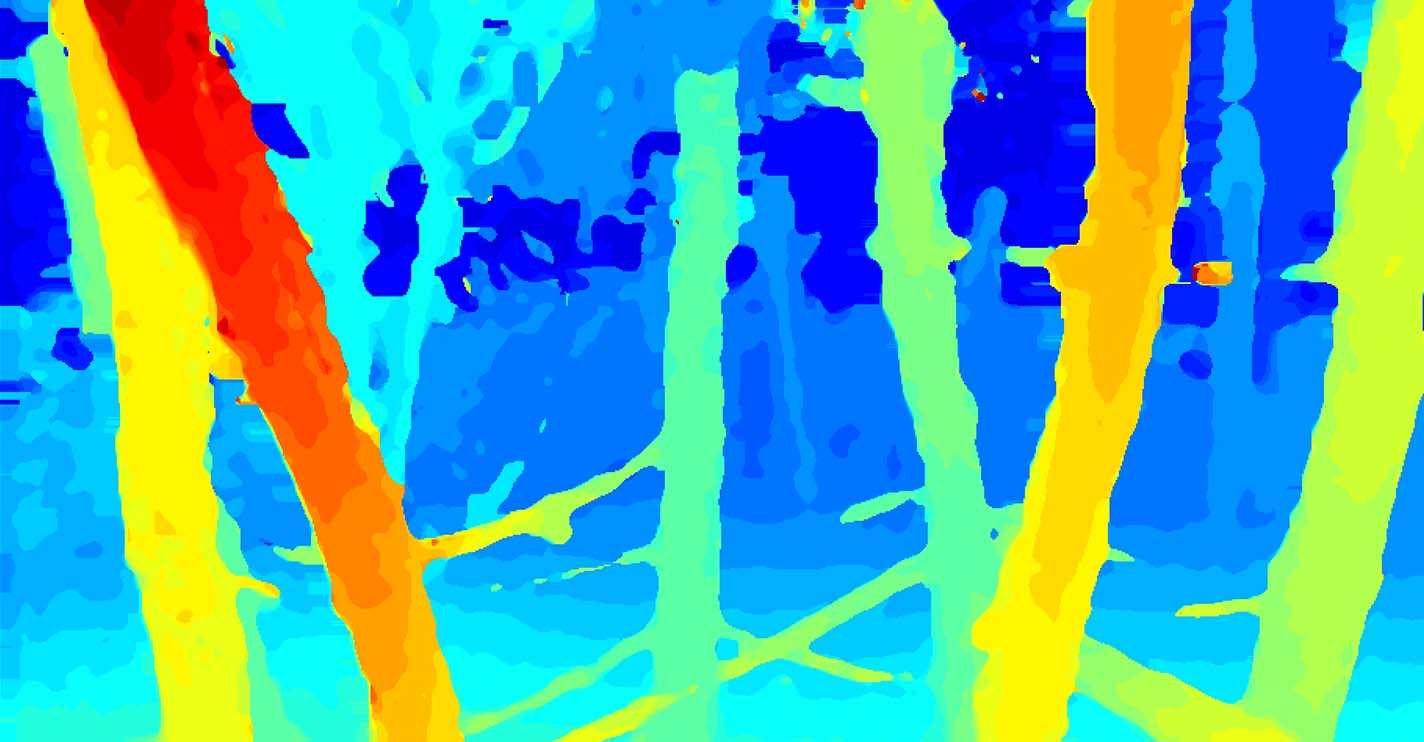}

\caption{Qualitative results from the ETH3D train and test dataset. The left column shows the first image of the stereo pair, the right column the final disparity map.}
\label{fig:results_eth3d_train}
\end{figure}

Fig.~\ref{fig:results_eth3d_train} shows some qualitative results of the ETH3D~\cite{eth3d} test and training set. It shows that while some details within the subpixel range might be missing, the overall structure is predicted very well.

\section{Generalization experiments}

In order to show that the network is not overfitted on a single dataset, we perform a simple generalization test, where the weight trained on one dataset is used in order to do inference on the other two datasets. The 2-point error is reported. Missing or invalid ground-truth measurements are not taken into account in this evaluation. This changes the metric for the KITTI2012~\cite{kitti} benchmark where the official benchmark interpolates the missing ground-truth measurements.

\begin{table}
\center
\caption{Generalization test}
\label{tab:gen_test}
\begin{tabular}{|c|c|c|c|}
\hline
& Middlebury~\cite{mb} & Kitti2012~\cite{kitti} & ETH3D~\cite{eth3d}\\ 
& (trained)    & (trained)   & (trained) \\
\hline
Middlebury~\cite{mb} & \textbf{17.9} & 19.6 & 20.4 \\
\hline
Kitti2012~\cite{kitti} & 20.30 & \textbf{16.66} & 19.70\\
\hline
ETH3D~\cite{eth3d} & 7.65 & 6.78 & \textbf{5.77}\\ 
\hline
\end{tabular}
\end{table}

Tab.~\ref{tab:gen_test} shows that while the best performance is achieved by training on the corresponding dataset, the network performance stays stable even when trained on completely different scenes. This shows that our method generalizes well and is usable for many different applications. However, the conducted experiments suggest that the KITTI2012~\cite{kitti} benchmark profits the most from training, as the achieved accuracies varied the most of all tested datasets.

\section{Conclusion and future work}

We have shown that a dense-network structure can be advantageous for the stereo vision task. By using this structure and by not using any fully-connected layers or 3D convolutions we were able to produce a very lightweight network that still has comparable results on difficult outdoor and indoor datasets.

Instead of relying on out-of-the-box post-processing solutions that are often not optimized for GPU usage such as semi-global matching or conditional random fields we use our own three-steps post processing. First we filter the cost-volume, then we detect inconsistencies that we afterwards update by using a foreground-background segmentation.

In the future we want to conduct an exhaustive network architecture study. To this end the number of layers experiments should be extended to also include number of feature maps. This may lead to an even better network structure.

The foreground-background segmentation may be improved upon by using more advanced techniques. For all of our experiments we hold the parameters for the segmentation of the disparity map static which will not lead to optimal solutions for each individual scene. This might be improved upon in the future by having an adaptive method or by using machine learning.

The runtime of our updating scheme for inconsistent points can be improved, especially when large portions of the image are inconsistent. This frequently happened in the KITTI benchmark, as often large portions of the image are sky, which cannot be correctly matched and therefore are marked as inconsistent.

In the future it might proof to be beneficial to learn the weights of the guided filter, instead of using pre-trained weights.






\begin{thebibliography}{1}


\bibitem{feat:census}
R. Zabih, et al. \emph{Non-parametric local transforms for computing visual correspondence.} European conference on computer vision. Springer, Berlin, Heidelberg, 1994.

\bibitem{feat:dense}
P. Pinggera, et al. \emph{On cross-spectral stereo matching using dense gradient features.}, Computer Vision and Pattern Recognition (CVPR), 2012 IEEE Conference on. Vol. 2, 2012.

\bibitem{disp:mc_cnn}
J. Zbontar and Y. LeCun. \emph{Stereo matching by training a convolutional neural network to compare image patches.} Journal of Machine Learning Research, Vol. 17, p. 1-32, 2016.

\bibitem{disp:cnn_crf}
P. Knobelreiter et al. \emph{End-to-end training of hybrid CNN-CRF models for stereo.} Proceedings of the IEEE Conference on Computer Vision and Pattern Recognition, 2017.

\bibitem{disp:ga_net}
F. Zhang, et al. \emph{Ga-net: Guided aggregation net for end-to-end stereo matching.} Proceedings of the IEEE Conference on Computer Vision and Pattern Recognition, 2019.

\bibitem{disp:psm_net}
J.R. Chang and Y.S. Chen. {Pyramid stereo matching network.} Proceedings of the IEEE Conference on Computer Vision and Pattern Recognition, 2018.

\bibitem{disp:gc_net}
A. Kendall, et al. \emph{End-to-end learning of geometry and context for deep stereo regression.} Proceedings of the IEEE International Conference on Computer Vision, 2017.

\bibitem{disp:efficient_stereo}
W. Luo, et al. \emph{Efficient deep learning for stereo matching.} Proceedings of the IEEE Conference on Computer Vision and Pattern Recognition, 2016.

\bibitem{densenet}
G. Huang, et al. \emph{Densely connected convolutional networks.} Proceedings of the IEEE conference on computer vision and pattern recognition, 2017.

\bibitem{feat:sadssd}
D. Kong and H. Tao. \emph{A Method for Learning Matching Errors in Stereo Computation.} BMVC, p. 2, 2004.

\bibitem{feat:ncc}
Y. S.Heo, et al. \emph{Robust stereo matching using adaptive normalized cross-correlation.} IEEE Transactions on pattern analysis and machine intelligence, 33(4), p. 807-822, 2010.

\bibitem{mi}
P. Viola and W.M. Wells III. \emph{Alignment by maximization of mutual information.} International journal of computer vision 24.2, p. 137-154, 1997.

\bibitem{reg:sgm}
 H. Hirschmueller. \emph{Accurate and efficient stereo processing by semi-global matching and mutual information.} IEEE Computer Society Conference on Computer Vision and Pattern Recognition (CVPR'05). Vol. 2., p. 807-814, 2005.

\bibitem{reg:mgm}
G. Facciolo, et al. \emph{MGM: A significantly more global matching for stereovision.} 2015.

\bibitem{kornia}
E. Riba, et al. \emph{Kornia: an open source differentiable computer vision library for pytorch.} The IEEE Winter Conference on Applications of Computer Vision, 2020.

\bibitem{guided_filter}
H. Wu, et al. \emph{Fast end-to-end trainable guided filter.} Proceedings of the IEEE Conference on Computer Vision and Pattern Recognition, 2018.

\bibitem{psmnet}
JR Chang and YS Chen. \emph{Pyramid stereo matching network.}, Proceedings of the IEEE Conference on Computer Vision and Pattern Recognition, 2018.

\bibitem{feat:eval_cost}
H. Hirschmuller, D. Scharstein. \emph{Evaluation of stereo matching costs on images with radiometric differences.} IEEE transactions on pattern analysis and machine intelligence 31.9, p. 1582-1599, 2008.

\bibitem{fast-rcnn}
S. Ren, et al. \emph{Faster r-cnn: Towards real-time object detection with region proposal networks.} Advances in neural information processing systems, 2015.

\bibitem{locnet}
S. Gidaris and N. Komodakis. \emph{Locnet: Improving localization accuracy for object detection.} Proceedings of the IEEE conference on computer vision and pattern recognition, 2016.

\bibitem{NLA}
F. Tosi, et al. \emph{Leveraging confident points for accurate depth refinement on embedded systems.} Proceedings of the IEEE Conference on Computer Vision and Pattern Recognition Workshops, 2019.

\bibitem{reg:drr}
S. Gidaris and N. Komodakis. \emph{Detect, replace, refine: Deep structured prediction for pixel wise labeling.} Proceedings of the IEEE conference on computer vision and pattern recognition, 2017.

\bibitem{reg:subpx}
D. Scharstein and R. Szeliski. \emph{A taxonomy and evaluation of dense two-frame stereo correspondence algorithms.} International journal of computer vision 47.1-3, p. 7-42, 2002.

\bibitem{reg:subpx_and_cons}
H. Hirschmueller et al. \emph{Real-time correlation-based stereo vision with reduced border errors.} International Journal of Computer Vision 47.1-3, p. 229-246 , 2002.

\bibitem{reg:cons_check1}
Q. Yang et al. \emph{Stereo matching with color-weighted correlation, hierarchical belief propagation, and occlusion handling.} IEEE Transactions on Pattern Analysis and Machine Intelligence 31.3, p. 492-504, 2008.

\bibitem{reg:cons_check2}
C. Lei et al. \emph{Region-tree based stereo using dynamic programming optimization.} 2006 IEEE Computer Society Conference on Computer Vision and Pattern Recognition (CVPR'06). Vol. 2. IEEE, 2006.

\bibitem{reg:int_gaps}
H. Hirschmueller. \emph{Stereo vision based mapping and immediate virtual walkthroughs.} 2003.


\bibitem{act:relu}
V. Nair and G.E. Hinton. \emph{Rectified linear units improve restricted boltzmann machines.} Proceedings of the 27th international conference on machine learning (ICML-10), 2010.

\bibitem{disp:tanh1}
C. Bailer, et al. \emph{CNN-based patch matching for optical flow with thresholded hinge embedding loss.} Proceedings of the IEEE Conference on Computer Vision and Pattern Recognition, 2017.

\bibitem{disp:tanh2}
M. Brown, et al. \emph{Discriminative learning of local image descriptors.} IEEE transactions on pattern analysis and machine intelligence 33.1, p. 43-57, 2010.

\bibitem{S_and_P}
R.H. Chan, et al. \emph{Salt-and-pepper noise removal by median-type noise detectors and detail-preserving regularization.} IEEE Transactions on image processing 14.10, p. 1479-1485, 2005.

\bibitem{pytorch}
A. Paszke, et al.\emph{PyTorch: An Imperative Style, High-Performance Deep Learning Library.} Advances in Neural Information Processing Systems 32,
p. 8024--8035, 2019.

\bibitem{adam}
D.P. Kingma and J.Ba. \emph{Adam: A method for stochastic optimization.} arXiv preprint arXiv:1412.6980, 2014.

\bibitem{kitti}
M. Menze, and A. Geiger. \emph{Object scene flow for autonomous vehicles.} Proceedings of the IEEE conference on computer vision and pattern recognition., 2015.

\bibitem{mb}
D. Scharstein, et al. \emph{High-resolution stereo datasets with subpixel-accurate ground truth.} German conference on pattern recognition. Springer, Cham, 2014.

\bibitem{opencv}
G. Bradski. \emph{Open Source Computer Vision Library.} 2015.

\bibitem{eth3d}
Thomas Sch\"ops et al. \emph{A Multi-View Stereo Benchmark with High-Resolution Images and Multi-Camera Videos.} CVPR, 2017.

\bibitem{disp:gf_census}
K. Zhang, et al.\emph{Cross-Scale Cost Aggregation for Stereo Matching.} CVPR, 2014.

\bibitem{disp:oasm}
L. Ang and Y. Zejian. 
\emph{Occlusion Aware Stereo Matching via Cooperative Unsupervised Learning.} Proceedings of the Asian Conference on Computer Vision, ACCV, 2018.

\bibitem{iresnet}
Z. Liang, et al. \emph{Learning for disparity estimation through feature constancy.}, Proceedings of the IEEE Conference on Computer Vision and Pattern Recognition 2018.

\bibitem{disp:adsm}
O. Zeglazi et al.
\emph{Accurate dense stereo matching
for road scenes.} IEEE International
Conference on Image Processing (ICIP), p. 720-724, 2017.

\bibitem{disp:mesh_stereo}
C. Zhang, et al. \emph{Meshstereo: A global stereo model with mesh alignment regularization for view interpolation.}, Proceedings of the IEEE International Conference on Computer Vision, 2015.

\bibitem{disp:elas}
G. Andreas, et al. \emph{Efficient large-scale stereo matching.}, Asian conference on computer vision. Springer, Berlin, Heidelberg, 2010.

\bibitem{sgm_mi}
H. Hirschmuller. \emph{Stereo processing by semiglobal matching and mutual information.}, IEEE Transactions on pattern analysis and machine intelligence 30.2, p. 328-341, 2007.

\bibitem{sgm_fie}
R. Schuster, et al. \emph{Combining stereo disparity and optical flow for basic scene flow.}, Commercial Vehicle Technology 2018, Springer Vieweg, Wiesbaden, p. 90-101, 2018.
\end{thebibliography}
%

\end{document}